\documentclass[online]{corr}
\usepackage[dvipsnames,svgnames]{xcolor} \usepackage[normalem]{ulem}

\makeatletter
\font\uwavefont=lasyb10 scaled 700
\def\spelling{\bgroup\markoverwith{\lower3.5\p@\hbox{\uwavefont\textcolor{Red}{\char58}}}\ULon}
\def\grammar{\bgroup\markoverwith{\lower3.5\p@\hbox{\uwavefont\textcolor{LimeGreen}{\char58}}}\ULon}
\def\phrasing{\bgroup\markoverwith{\lower3.5\p@\hbox{\uwavefont\textcolor{RoyalBlue}{\char58}}}\ULon}

\newcommand\remove{\bgroup\markoverwith{\textcolor{red}{\rule[0.5ex]{2pt}{0.4pt}}}\ULon}
\makeatother 
\usepackage{amsmath,amssymb}
\usepackage{cancel}
\usepackage{graphicx}
\usepackage{multirow}
\usepackage{enumitem}

\usepackage{natbib}

\usepackage[thmmarks,amsmath,framed]{ntheorem}
\theoremstyle{plain}
\theoremheaderfont{\normalfont\bfseries}
\theorembodyfont{\itshape}
\newtheorem{theorem}{Theorem}
\newtheorem{lemma}[theorem]{Lemma}

\newtheorem{corollary}[theorem]{Corollary}
\theoremstyle{plain}
\theoremheaderfont{\bfseries}
\theorembodyfont{\upshape}

\theoremstyle{break}
\newtheorem{definition}[theorem]{Definition}
\theoremstyle{nonumberplain}
\theoremheaderfont{\scshape}
\theorembodyfont{\upshape}
\theoremseparator{:}
\theoremsymbol{\hfill\ensuremath{_\Box}}
\renewtheorem{proof}{Proof}

\usepackage{cleveref}
\usepackage{fancyvrb}
\usepackage{relsize}
\usepackage{listings}
\usepackage{verbatim}
\usepackage{alltt}
\usepackage{booktabs}
\usepackage[group-separator={,}]{siunitx}
\hypersetup{
  pdftitle={Notation3 as an Existential Rule Language},
  pdfauthor={Dörthe Arndt and Stephan Mennicke},
  pdfkeywords={Notation3, RDF, Blank Nodes, Existential Rules},
	pdfsubject={Under consideration in Theory and Practice of Logic Programming (TPLP)},
}

\RequirePackage{xspace}

\newcommand\Slang[1]{\ensuremath{\mathbf{#1}}\xspace}
\newcommand\constS{\Slang{C}}
\newcommand\nullS{\Slang{N}}
\newcommand\varS{\Slang{V}}
\newcommand\predS{\Slang{P}}
\newcommand\arity[1]{\mathit{ar}(#1)\xspace}
\newcommand\Inst[1][I]{\ensuremath{\mathcal{#1}}\xspace}
\newcommand\Dnst{\Inst[D]}
\newcommand\arule[1][]{\ensuremath{r_{#1}}\xspace}
\newcommand\ruleS[1][\Sigma]{\ensuremath{#1}\xspace}

\newcommand\Requiv{\ensuremath{\mathrel{\leftrightarrows}}}
\newcommand\nthree{{N\oldstylenums 3}\xspace}
\newcommand\nex{$\nthree^\exists$\xspace}
\newcommand\comp{\ensuremath{\operatorname{comp}}}

\renewcommand{\vec}[1]{\mathbf{#1}}

\newcommand\transf[1][T]{\ensuremath{\mathcal #1}\xspace}

\newcommand\Ntequiv{\ensuremath{\mathrel{\equiv}}}
\newcommand\uvar[1]{\ensuremath{v_{\texttt{#1}}^{\forall}}}
\newcommand\evar[1]{\ensuremath{v_{\texttt{#1}}^{\exists}}}

\newcommand\listingsize{\fontsize{9pt}{10pt}}
\RecustomVerbatimCommand{\Verb}{Verb}{fontsize=\listingsize}
\RecustomVerbatimEnvironment{Verbatim}{Verbatim}{fontsize=\listingsize}
\definecolor{grey}{RGB}{130,130,130}
\definecolor{lightgrey}{RGB}{170, 170, 170}
\usepackage{tikz}
\usetikzlibrary{shapes.geometric}

\tikzset{
	object/.style = {draw=black, ellipse},
	literal/.style = {draw=black, rectangle},
	predicate/.style = {fill=white}
}

\begin{document}

\lefttitle{Dörthe Arndt and Stephan Mennicke}

\jnlPage{1}{8}
\jnlDoiYr{2021}
\doival{10.1017/xxxxx}

\title[Existential Notation3 Logic]{Existential Notation3 Logic\thanks{This work is partly supported by 
Deutsche Forschungsgemeinschaft (DFG, German Research Foundation) in project number 389792660 (TRR 248, \href{https://www.perspicuous-computing.science/}{Center for 	Perspicuous Systems}),
by the Bundesministerium für Bildung und Forschung (BMBF, Federal
Ministry of Education and Research) in project number 13GW0552B (\href{https://digitalhealth.tu-dresden.de/projects/kimeds/}{KIMEDS}),
in the \href{https://www.scads.de/}{Center for Scalable Data Analytics and
Artificial Intelligence} (ScaDS.AI),
and by BMBF and DAAD (German Academic Exchange Service) in project 57616814 (SECAI, School of Embedded and Composite AI).}
}

\begin{authgrp}
\author{\sn{Dörthe} \gn{Arndt}}
\affiliation{Computational Logic Group, TU Dresden, Germany}
\affiliation{ScaDS.AI, Dresden/Leipzig, Germany}
\author{\sn{Stephan} \gn{Mennicke}}
\affiliation{Knowledge-Based Systems Group, TU Dresden, Germany}
\end{authgrp}

\maketitle

\begin{abstract}
	In this paper, we delve into Notation3 Logic (N3), an extension of RDF, which empowers users to craft rules introducing fresh blank nodes to RDF graphs. This capability is pivotal in various applications such as ontology mapping, given the ubiquitous presence of blank nodes directly or in auxiliary constructs across the Web. However, the availability of fast N3 reasoners fully supporting blank node introduction remains limited. Conversely, engines like VLog or Nemo, though not explicitly designed for Semantic Web rule formats, cater to analogous constructs, namely existential rules.

	We investigate the correlation between N3 rules featuring blank nodes in their heads and existential rules. We pinpoint a subset of N3 that seamlessly translates to existential rules and establish a mapping preserving the equivalence of N3 formulae. To showcase the potential benefits of this translation in N3 reasoning, we implement this mapping and compare the performance of N3 reasoners like EYE and cwm against VLog and Nemo, both on native N3 rules and their translated counterparts. Our findings reveal that existential rule reasoners excel in scenarios with abundant facts, while the EYE reasoner demonstrates exceptional speed in managing a high volume of dependent rules.
	
	Additionally to the original conference version of this paper, we include all proofs of the theorems and introduce a new section dedicated to N3 lists featuring built-in functions and how they are implemented in existential rules.
	Adding lists to our translation/framework gives interesting insights on related design decisions influencing the standardization of N3.
\end{abstract}

\section{Introduction}\label{sec:intro}

Notation3 Logic (\nthree) is an extension of the Resource Description Framework (RDF) which allows the user to quote graphs, to express rules, and to apply built-in functions on the components of RDF triples \citep{N3-spec,BernersLeeCKH08:N3Logic:tplp}.
Facilitated by reasoners like cwm \citep{cwm}, Data-Fu \citep{datafu}, or EYE \citep{eyepaper},
\nthree rules directly consume and produce RDF graphs. This makes \nthree well-suited for rule exchange on the Web.
\nthree supports the introduction of new blank nodes through rules, that is, if a blank node appears in the head\footnote{To stay consistent  across frameworks, we use the terms \emph{head} and \emph{body} throughout the whole paper. 
The head is the part of the rule occurring 
at the end of the implication arrow, the body the part at its beginning (backward rules:
``$\text{head}\leftarrow \text{body}$'', forward rules: ``$\text{body}\rightarrow \text{head}$'').} of a rule,
each new match for the rule body produces a new instance of the rule's head containing \emph{fresh} blank nodes.  This feature is interesting for many use cases -- mappings between different vocabularies include blank nodes,
workflow composition deals with unknown existing instances \citep{Arndt17:pragmaticProof:tplp} -- but it also impedes reasoning tasks: from a logical point of view these rules contain existentially quantified variables in their heads. Reasoning with such rules is known to be undecidable in general  and very complex on decidable cases \citep{BLMS11:decline,KMR:ChasePower2019}.

   Even though recent projects like jen3\footnote{\url{https://github.com/william-vw/jen3}} or RoXi \citep{BO23:roxi} aim at improving the situation,
the number of fast \nthree reasoners 
fully  supporting blank node introduction 
   is low. This is different for reasoners
acting on  existential rules,
a concept 
 very similar to  blank-node-producing rules in \nthree, but developed for databases.
Sometimes it is necessary to uniquely identify data by a value that is not already part of the target database.
   One tool to achieve that are \emph{labeled nulls} which -- just as blank nodes -- indicate \emph{the existence} of a value.
This problem from databases and the observation that rules may provide a powerful, yet declarative, means of computing has led to more extensive studies of existential rules \citep{BLMS11:decline,gottlobrr}.
Many reasoners like for example VLog \citep{VLog4j2019} or Nemo \citep{nemo23} apply dedicated strategies to optimize reasoning with existential rules.

This paper aims to make existing and future optimizations on existential rules usable in the Semantic Web.
We introduce a subset of \nthree supporting existential quantification but ignoring features of the language not 
covered in existential rules, like for example built-in functions or lists.
We provide a mapping between this logic and existential rules:
The mapping and its inverse both preserve equivalences of formulae, enabling \nthree reasoning via existential rule technologies.
We discuss how the framework can be extended to also support lists -- a feature of \nthree used in many practical applications, for example to support n-ary predicates.
We implement the defined mapping in python
 and compare
 the reasoning performance of the existential rule reasoners Vlog and Nemo, and the \nthree reasoners EYE and cwm for two benchmarks:
one applying a fixed set of rules on a varying size of facts, and
one applying a varying set of highly dependent rules to a fixed set of facts.
In our tests VLog and Nemo together with our mapping outperform the traditional \nthree reasoners EYE and cwm
 when dealing with a high number of facts while EYE is the fastest on large dependent rule sets.
This is a strong indication that our implementation will be of practical use
when extended by further features. 

We motivate our approach by providing examples of \nthree and existential rule formulae, and discuss how these are connected, in \Cref{yea!}.
In \Cref{n3} we provide a more formal definition of Existential \nthree (\nex), introduce its semantics and discuss its properties.
We then formally introduce existential rules, provide the mapping from \nex into this logic, and prove its truth-preserving properties in \Cref{exru}. \nthree lists and the built-ins associated with them are introduced as \nthree primitives as well as their existential rule translations are subject to \Cref{sec:lists}.
In \Cref{imp} we discuss our implementation and provide an evaluation of the different reasoners.
Related work is presented in \Cref{relwork}.
We conclude our discussion in \Cref{conc}.
Furthermore, the code needed for reproducing our experiments is available on \href{https://github.com/smennicke/n32rules}{GitHub} (\url{https://github.com/smennicke/n32rules}).

This article is an extended and revised version of our work \citep{rr23n3} presented at Rules and Reasoning -- 7th International Joint Conference (RuleML+RR) 2023.
Compared to the conference paper, we include full proofs to all theorems and lemmas.
Furthermore, we strengthen the statements of correctness of our translation (\Cref{thm:full-abstraction} in \Cref{exru}), imposing stronger guarantees with effectively the same proofs as we had for the conference version, back then included in the technical appendix \citep{arxiv2023report} only.
A discussion about the particular difference is appended to \Cref{thm:full-abstraction}.
Finally, we extend our considerations by \nthree lists and respective built-ins (cf.\ \Cref{sec:lists}).

 \section{Motivation}\label{yea!}
\setlength{\abovedisplayskip}{.25em}
\setlength{\belowdisplayskip}{.25em}
\nthree has been inroduced  as a rule-based extension of RDF.
As in RDF, \nthree knowledge is stated in triples consisting of \emph{subject}, \emph{predicate}, and \emph{object}.  In ground triples these can either be Internationalized Resource Identifiers (IRIs) or literals. The expression
\begin{equation}\label{eq:fact-n3}
	\texttt{:lucy :knows :tom.}
\end{equation}
means\footnote{We omit name spaces for brevity.} that \emph{``lucy knows tom''}. Sets of triples are interpreted as their conjunction.
Like RDF, 
\nthree supports blank nodes, usually starting with \texttt{\_:}, which stand for (implicitly) existentially quantified variables. 
The statement
\begin{equation}\label{eq:blank-n3}
	\texttt{:lucy :knows \_:x.}
\end{equation}
means \emph{``there exists someone who is known by lucy''}. \nthree furthermore supports implicitly universally quantified variables, indicated by a leading question mark (\texttt{?}), and implications
which are stated using graphs, i.e., sets of triples, surrounded by curly braces (\texttt{\{\}}) as body and head connected via an arrow (\texttt{=>}). The formula
\begin{equation}\label{eq:dlog-n3}
	\texttt{\{:lucy :knows ?x\}=>\{?x  :knows :lucy\}.}
\end{equation}
means that \emph{``everyone known by Lucy also knows her''}.
Furthermore, \nthree allows the use of blank nodes in rules.
These blank nodes are not quantified outside the rule like the universal variables, but in the rule part they occur in, that is either in its  body or its head.
\begin{equation}\label{eq:exq-n3}
	\texttt{\{?x :knows  :tom\}=>\{?x :knows \_:y. \_:y :name "Tom"\}.}
\end{equation}
means \emph{``everyone knowing Tom  knows \emph{someone} whose name is \emph{Tom}''}.

\newcommand\tripl{\ensuremath{\textit{tr}}\xspace}
\newcommand\ntt[1]{\texttt{#1}\xspace}
This last example shows, that \nthree supports rules concluding the \emph{existence} of certain
 terms 
which makes it easy to express them as  \emph{existential rules}.
An existential rule is a first-order sentence of the form
\begin{equation}\label{eq:exru-shape}
    \forall \vec{x}, \vec{y} .\,\varphi[\vec x, \vec y] \rightarrow \exists \vec{z} .\,\psi[\vec y, \vec z]
\end{equation}
where $\vec x, \vec y, \vec z$ are mutually disjoint lists of variables, $\varphi$ and $\psi$ are conjunctions of atoms using only variables from the given lists, and $\varphi$ is referred to as the \emph{body} of the rule while $\psi$ is called the \emph{head}.
Using the basic syntactic shape of \eqref{eq:exru-shape} we go through all the example \nthree formulae \eqref{eq:fact-n3}--\eqref{eq:exq-n3} again and represent them as existential rules.
To allow for the full flexibility of \nthree and RDF triples, we translate each RDF triple, just like the one in \eqref{eq:fact-n3} into a first-order atom $\tripl(\texttt{:lucy}, \texttt{:knows}, \texttt{:tom})$.
Here, \tripl\xspace is a ternary predicate holding subject, predicate, and object of a given RDF triple.
This standard translation makes triple predicates (e.g., $\ntt{:knows}$) accessible as terms.
First-order atoms are also known as \emph{facts}, finite sets of facts are called \emph{databases}, and (possibly infinite) sets of facts are called \emph{instances}.
Existential rules are evaluated over instances (cf.\xspace \Cref{exru}).

Compared to other rule languages, the distinguishing feature of existential rules is the use of existentially quantified variables in the head of rules (cf.\xspace $\vec z$ in~\eqref{eq:exru-shape}).
The \nthree formula in~\eqref{eq:blank-n3} contains an existentially quantified variable and 
can, thus, 
be encoded as \begin{equation}
  \label{eq:blank-rl}
  \rightarrow \exists x.\ \tripl(\ntt{:lucy}, \ntt{:knows}, x)
\end{equation}\sloppy
Rule~\eqref{eq:blank-rl} has an empty body, which means the head is unconditionally true. Rule \eqref{eq:blank-rl} is satisfied on instances containing any fact $\tripl(\ntt{:lucy},\ntt{:knows},\_)$ (e.g., $\tripl(\ntt{:lucy},\ntt{:knows},\ntt{:tim})$ so that variable $x$ can be bound to $\ntt{:tim}$).

The implication of~\eqref{eq:dlog-n3} has
\begin{equation}
  \label{eq:dlog-rl}
  \forall x .\ \tripl(\ntt{:lucy},\ntt{:knows},x) \rightarrow \tripl(x, \ntt{:knows}, \ntt{:lucy})
\end{equation}
as its (existential) rule counterpart, which does not contain any existentially quantified variables. Rule~\eqref{eq:dlog-rl} is satisfied in the instance
$$\Inst_{1} = \{ \tripl(\texttt{:lucy}, \texttt{:knows}, \ntt{:tom}), \tripl(\ntt{:tom}, \ntt{:knows},\ntt{:lucy}) \}$$
but not in
$$\Inst[K]_{1} = \{ \tripl(\ntt{:lucy}, \ntt{:knows}, \ntt{:tom}) \}$$
since the only fact in $\Inst[K]_1$ matches the body of the rule, but there is no fact reflecting on its (instantiated) head (i.e., the required fact $\tripl(\ntt{:tom},\ntt{:knows},\ntt{:lucy})$ is missing). 
Ultimately, the implication \eqref{eq:exq-n3} with blank nodes in its head may be transferred to a rule with an existential quantifier in the head:
\begin{equation}
  \label{eq:exq-rl}
  \forall x.\ \tripl(x, \ntt{:knows}, \ntt{:tom}) \rightarrow \exists y .\ \left(\tripl(x, \ntt{:knows}, y) \wedge \tripl(y, \ntt{:name}, \ntt{"Tom"})\right)\text.
\end{equation}
It is clear that rule~\eqref{eq:exq-rl} is satisfied in instance
$$\Inst_{2} = \{ \tripl(\ntt{:lucy},\ntt{:knows},\ntt{:tom}), \tripl(\ntt{:tom}, \ntt{:name}, \ntt{"Tom"}) \}\text.$$
However, instance $\Inst[K]_1$ does not satisfy rule~\eqref{eq:exq-rl} because although the only fact satisfies the rule's body, there are no facts jointly satisfying the rule's head.

Note, for query answering over databases and rules, it is usually not required to decide for a concrete value of $y$ (in rule~\eqref{eq:exq-rl}).
Many implementations, therefore, use some form of abstraction: for instance, Skolem terms.
VLog and Nemo implement the \emph{standard chase} which uses another set of terms, so-called \emph{labeled nulls}.
Instead of injecting arbitrary constants for existentially quantified variables, (globally) fresh nulls are inserted in the positions existentially quantified variables occur.
Such a labeled null embodies the existence of a constant on the level of instances (just like blank nodes in RDF graphs).
Let $n$ be such a labeled null.
Then $\Inst_{2}$ can be generalized to
$$\Inst_{3} = \{ \tripl(\ntt{:lucy},\ntt{:knows},\ntt{:tom}), \tripl(\ntt{:lucy},\ntt{:knows},n), \tripl(n, \ntt{:name}, \ntt{"Tom"}) \}\text,$$
on which rule~\eqref{eq:exq-rl} is satisfied, binding null $n$ to variable $y$.
$\Inst_{3}$ is, in fact, more general than $\Inst_{2}$ by the following observation:
There is a mapping from $\Inst_{3}$ to $\Inst_{2}$ that is a homomorphism (see \Cref{sub:rule-prelims} for a formal introduction) but not vice versa.
The homomorphism here maps the null $n$ (from $\Inst_{3}$) to the constant $\ntt{:tom}$ (in $\Inst_{2}$).
Intuitively, the existence of a query answer (for a conjunctive query) on $\Inst_{3}$ implies the existence of a query answer on $\Inst_{2}$.
Existential rule reasoners implementing some form of \emph{the chase} aim at finding the most general instances (\emph{universal models}) in this respect \citep{DNR08:corechase}.

In the remainder of this paper, we further analyze the relation between \nthree and existential rules.
First, we give a brief formal account of the two languages and then provide a correct translation function from \nthree to existential rules.

\section{Existential N3}\label{n3}
In the previous section we introduced essential elements of \nthree, namely
triples and rules. 
\nthree also supports more complex constructs like lists, nesting of rules, and quotation.
As these features are not covered by existential rules, we define a subset of \nthree excluding them, called \emph{existential \nthree} (\nex). 
This fragment of \nthree is still very powerful as it covers ontology mapping, one of N3's main use cases. Many ontologies rely on patterns including auxiliary blank nodes. \nex supports the production of these.\footnote{A very good example of such an ontology is the RDF version of HL7 FHIR (\url{https://www.hl7.org/fhir/rdf.html}). In HL7 FHIR literal values are always used in combination with the predicate \texttt{fhir:v}. The connection to concepts is always done through blank nodes representing so-called \emph{primitive elements}. If we want to map from ontologies containing various datatype properties like for example FOAF (\url{http://xmlns.com/foaf/spec/}) to FHIR, we need to create new blank nodes.} In practice, these mappings are often connected with build-in functions like calculations or string operations\footnote{Spllitting the first name from the last name in a string if the target ontology requires to have these separated would be a practical example.}, these are not covered yet, but could be added. A more difficult feature to add would be the support of so-called  rule-producing rules: In N3 it is possible to nest rules into the head of other rules. While this technique does not yield more expressivity, it is commonly used to translate from RDF datasets to N3 rules (see e.g., \cite{rprs}).  Such rule-producing rules can not be coverd by existential rules as these only allow the derivation of facts.

We base our definitions on
so-called \emph{simple \nthree formulae} \citep[Chapter 7]{Arndt19:PhD}, 
these are \nthree formulae which do not allow for nesting.

\subsection{Syntax}
\nex relies on the RDF alphabet. As the distinction is not relevant in our context, we consider IRIs and literals together as constants.
Let  $C$ be a set of such constants, $U$ a set of universal variables  (starting with \texttt{?}), and  $E$ a set of existential variables (i.e., blank nodes).
If the sets $C$, $U$, $E$, and $\{\texttt{\{}, \texttt{\}}, \texttt{=>},  \texttt{.}\}$ are mutually disjoint, we call
$\mathfrak{A}:=C \,\cup\, U \,\cup\, E \,\cup\, \{\texttt{\{}, \texttt{\}}, \texttt{=>},  \texttt{.}\}$ an \emph{\nthree alphabet}.
\Cref{N3S} provides the syntax of $\nthree^\exists$ over $\mathfrak{A}$.

\begin{figure}[tbp]
	\begin{tabular}{lp{0.15\textwidth}rp{0.02 \textwidth}|lp{0.1 \textwidth}r}
		\hline
\texttt{f ::= } & &                   formulae:&& \texttt{t ::=}&&                    terms:\\ 
		&  \texttt{t t t.}&                atomic formula&&  & \texttt{ex} &                existential variables\\
		&  \texttt{\{e\}=>\{e\}.}& implication   & & & \texttt{c} &                constants\\
&  \texttt{f f} &                 conjunction&\\
		&&&&\\
		\texttt{n ::=}&&                    N3 terms:& & \texttt{e ::=}&&                    expressions:\\
		& \texttt{uv} &                universal variables & &&\texttt{n n n.} &               triple expression\\
		&         \texttt{t} &               terms  &&&\texttt{e e} &               conjunction expression\\
		\hline
	\end{tabular}
	\caption{Syntax of $\operatorname{N3}^\exists$}\label{N3S}
\end{figure}

\nex fully covers RDF. RDF formulae are conjunctions of atomic formulae.  Just as generalized RDF \citep{rdf11-concepts}, \nex  allows for  literals and blank nodes to occur in subject, predicate, and object position. The same holds for universal variables which are not present in RDF.  This syntactical freedom is inherited from full N3 and makes it possible to -- among other things -- express the rules for RDF/S  \citep[Appendix A]{rdf11-semantics} and OWL-RL \citep[Section 4.3]{owl2-profiles}
entailment via N3.  As an example for that, consider the following rule\footnote{This rule corresponds to  \texttt{prp-inv1} in  OWL profiles \citep{owl2-profiles}.} for inverse properties:
\begin{equation}\label{variablepredicate}
	\texttt{\{?p1 owl:inverseOf ?p2 . ?x ?p1 ?y .\}=>\{?y ?p2 ?x \}.}
\end{equation}
If we apply this rule on triple~\eqref{eq:fact-n3} in combination with
\begin{equation}\label{eq:inverse}
	\texttt{:knows owl:inverseOf :isKnownBy.}
\end{equation}
we derive
\begin{equation}
	\texttt{:tom :isKnownBy :lucy.}
\end{equation}
Similar statements and rules can be made for triples including literals. We can for example declare that the \texttt{:name} from rule~\eqref{eq:exq-n3} is the \texttt{owl:inverseOf} of \texttt{:isNameOf}\footnote{Note that this is not possible in OWL itself as datatype properties cannot be inversed.}. With rule~\eqref{eq:inverse} we then derive from
\begin{equation}
	\texttt{\_:x :name "Tom".} 
\end{equation}
that
\begin{equation}
  \texttt{"Tom" :isNameOf \_:x.}
\end{equation}
In that sense the use of generalized RDF ensures that all logical consequences we are able to produce via rules can also be stated in the language.  This \emph{principle of syntactical completeness} is also the reason to allow literals and blank nodes in predicate position. As  universals may occur in  predicate position, this also needs to be the case for all other kinds of symbols.

Currently, there is one exception to our principle:  The syntax above allows
rules having new universal variables in their head like for example
\begin{equation}
	\texttt{\{:lucy :knows :tom\}=>\{?x :is :happy\}.}
\end{equation}
which results in a rule expressing \emph{``if lucy knows tom, everyone is happy''}. This implication is problematic: Applied on triple~\eqref{eq:fact-n3}, it yields $\texttt{?x :is :happy.}$ 
which is a triple containing a universal variable. Such triples are not covered by our syntax, the rule thus introduces a 
fact we cannot express. 
Therefore, we restrict \nex rules to \emph{well-formed implications} which rely on \emph{components}. \emph{Components}\label{formulacomponents} can be seen as direct  parts\footnote{As full N3 supports graph terms, it could also be seen as a subset of terms as defined for full N3.} an N3 formula consists of.
Let $f$ be a formula or an expression over an alphabet $\mathfrak{A}$.
The set $\comp(f)$ of \emph{components} of $f$ is defined as:
\begin{itemize}
	\item If $f$ is an atomic formula or a triple expression of the form $t_1~ t_2~ t_3.$, $\comp(f)=\{t_1,t_2,t_3\}$.
	\item If $f$ is an implication of the form $\verb!{!e_{1}\verb!}=>{!e_{2}\verb!}!.$, then $\comp(f)=\{\verb!{!e_1\verb!}!, \verb!{!e_2\verb!}!\}$.
	\item If $f$ is a conjunction of the form $f_1 f_2$, then $\comp(f)=\comp(f_1)\cup \comp(f_2)$.
\end{itemize}

A rule  $\texttt{\{e}_1\texttt{\}=>\{e}_2\texttt{\}} .$  is called  \emph{well-formed}  if $(\comp(\texttt{e}_2)\setminus \comp(\texttt{e}_1))\cap U=\emptyset$.
 For the remainder of this paper we assume all implications to be well-formed. Note that this definition of well-formed formulae is closely related to the idea of safety in logic programming.  Well-formed rules are safe.

\subsection{Semantics}\label{n3semantics}
In order to define the semantics of \nex we first note, that in our fragment of \nthree all quantification of variables is only defined implicitly.
The blank node in triple~\eqref{eq:blank-n3}  is understood as an existentially quantified variable, the universal in formula~\eqref{eq:dlog-n3} as universally quantified.  Universal quantification spans over the whole formula -- variable \texttt{?x}  occurring  in body and head of rule~\eqref{eq:dlog-n3} is universally quantified for the whole implication -- while existential quantification is local -- the conjunction in the head of rule~\eqref{eq:exq-n3} is existentially quantified there. Adding new triples as conjuncts to formula~\eqref{eq:exq-n3} like \begin{equation}\label{eq:2bl}
	\texttt{:lucy :knows \_:y. \_:y :likes :cake. }
\end{equation}	
leads to  the new statement that \emph{``lucy knows someone who likes cake''} but even though we are using the same blank node identifier \texttt{\_:y} in both formulae, the quantification of the variables in this formula is totally seperated and the person named ``Tom'' is not necessarily related to the cake-liker.  With the goal to deal with this  locality  of blank node scoping, we define substitutions which are only applied on components of formulae and leave nested elements like for example the body and head of rule~\eqref{eq:dlog-n3} untouched.

A \emph{substitution} $\sigma$
 is a mapping from a set of variables $X\subset U\cup E$ to the set of \nthree terms. We \emph{apply} $\sigma$ to a term, formula or expression $x$ as follows:
\begin{itemize}
	\item $x \sigma = \sigma(x)$ if $x\in X$,
	\item $(s~p~o) \sigma =(s\sigma)( p\sigma) (o\sigma)$ if $x=s~p~ o$ is an atomic formula
	or a triple expression,
	\item $(f_1f_2) \sigma=(f_1\sigma)( f_2\sigma)$ if $x=f_1f_2$ is a conjunction,
	\item $x \sigma = x$ else.\end{itemize}

For formula 
  $f=\texttt{\_:x :p :o. \{\_:x :b :c\}=>\{\_:x :d :e\}.}$, substitution  $\sigma$ and $\texttt{\_:x}\in \text{dom}(\sigma)$, we get:  $f\sigma=\sigma(\texttt{\_:x}) \texttt{:p :o. \{\_:x :b :c\}=>\{\_:x :d :e\}}$.\footnote{Note that
the semantics of \emph{simple formulae} on which \nex's semantics is based, relies
on two ways to apply a substitution which is necessary to handle nested rules, since such constructs are excluded in \nex, we simplified here.}
We use the substitution to define the semantics of \nex which additionally makes use of  \emph{N3 interpretations} $\mathfrak{I} = (\mathfrak{D},\mathfrak{a},\mathfrak{p})$
consisting of
(1) a set  $\mathfrak{D}$, called  the domain of $\mathfrak{I}$;
(2) a mapping $\mathfrak{a}: 
C\rightarrow \mathfrak{D}$, called the object function;
(3) a mapping $\mathfrak{p}:
\mathfrak{D} \rightarrow 2^{\mathfrak{D} \times \mathfrak{D}}$, called the predicate function.

Just as the function IEXT in RDF's simple interpretations \citep{rdf-semantics}, \nthree{}'s predicate function 
maps  elements from the domain of discourse to a set of pairs of domain elements and is not applied on relation symbols directly.
This makes quantification over predicates possible while not exceeding first-order logic in terms of complexity.  To  introduce the \emph{semantics of \nex}\label{sem_n3},
let $\mathfrak{I}=(\mathfrak{D},\mathfrak{a,p})$ be an \nthree interpretation. For an   \nex formula  $f$: \begin{enumerate}
\item \label{quant2} If $W=\text{comp}(f)\cap E \neq \emptyset$, then $\mathfrak{I}\models f$
 iff $\mathfrak{I}\models f\mu$ for some substitution $\mu: W\rightarrow C$.
  \item If  $\text{comp}(f)\cap E=\emptyset$:
  \begin{enumerate}
    \item If $f$ is an atomic formula $t_1\, t_2\, t_3$, then  $\mathfrak{I} \models t_1\, t_2\, t_3$.     iff $(\mathfrak{a}(t_1),\mathfrak{a}(t_3))\in\mathfrak{p}(\mathfrak{a}(t_2))$.
    \item If $f$ is a conjunction $f_1f_2$, then  $\mathfrak{I}\models f_1 f_2$ iff       $\mathfrak{I}\models f_1$ and $\mathfrak{I}\models f_2$.\label{conj}
    \item If $f$ is an implication\label{implication}, then $\mathfrak{I} \models \verb!{! e_1 \verb!}! \verb!=>! \verb!{! e_2 \verb!}!$ iff  $\mathfrak{I} \models e_2\sigma$ if $\mathfrak{I} \models e_1\sigma$
        for all substitutions $\sigma$ on the universal variables $\text{comp}(\texttt{e}_1)\cap U$ by constants.
    \end{enumerate}
\end{enumerate}

The semantics as defined above uses a substitution into the set of constants instead of a direct assignment to the domain of discourse to interpret quantified variables. This 
design choice inherited from \nthree ensures referential opacity of quoted graphs
 and means, in essence, that  quantification always refers to named domain elements.

With that semantics, we call
an interpretation $\mathfrak{M}$  \emph{model} of a dataset $\Phi$, written as $\mathfrak{M}\models\Phi$, if $\mathfrak{M}\models f$ for each formula $f\in\Phi$.
We say that two sets of \nex formulae
 $\Phi$ and $\Psi$ are \emph{equivalent}, written as $\Phi\equiv \Psi$, if for all interpretations $\mathfrak{M}$: $\mathfrak{M}\models \Phi$ iff $\mathfrak{M}\models \Psi$.  
If 
$\Phi=\{\phi\}$ and $\Psi=\{\psi\}$ are singleton sets, we write $\phi\equiv \psi$ omitting the brackets.

\paragraph{Piece Normal Form}
$\nthree{}^\exists$ formulae consist of conjunctions of triples and implications. 
For our goal of translating such formulae to existential rules, it is convenient to consider sub-formulae seperately.

Below, we  therefore define the so-called \emph{Piece Normal Form} (PNF) for $\nthree{}^\exists$ formulae and show that  each such formula $f$ is equivalent to a set of sub-formulae $\Phi$  (i.e., $\Phi \equiv \phi$) in PNF.
We proceed in two steps. 
First, we separate formulae based on their blank node components.  If two parts
of a conjunction share a blank node component, as in formula~\eqref{eq:2bl}, we cannot split the formula into two since the information about the co-reference would get lost. 
However, if conjuncts either do not contain blank nodes or only contain disjoint sets of these, we can split them into so-called \emph{pieces}:
Two formulae $f_1$ and $f_2$ are called \emph{pieces} of a formula $f$ if $f=f_1f_2$ and $\comp(f_1)\cap\comp(f_2)\cap E=\emptyset$. For such formulae  we know:
\begin{lemma}[Pieces]\label{lemma:pieces}
	Let $f=f_1f_2$ be an \nex conjunction and let $\comp(f_1)\cap\comp(f_2)\cap E=\emptyset$, then for each interpretation $\mathfrak{I}$,
	$\mathfrak{I}\models f \text{ iff } \mathfrak{I}\models f_1 \text{ and } \mathfrak{I}\models f_2$.
 
\end{lemma}
\begin{proof}
	 
	\begin{enumerate}
\item \label{c2} 
	If  $\comp(f)\cap E=\emptyset$ the claim follows immediately by point 2b in the semantics definition.
	\item If $W=\text{comp}(f)\cap E \neq \emptyset$:
	
	 ($\Rightarrow$)
			If $\mathfrak{I}\models f$ then there exists a substitution $\mu:\comp(f)\cap E\rightarrow C$ such that $\mathfrak{I}\models f\mu$, that is  $\mathfrak{I}\models (f_1\mu)~(f_2\mu)$. According to the previous point that implies
$\mathfrak{I}\models f_1\mu$ and $\mathfrak{I}\models f_2\mu$ and thus $\mathfrak{I}\models f_1$ and $\mathfrak{I}\models f_2$.
	
			($\Leftarrow$)
			If $\mathfrak{I}\models f_1$ and $\mathfrak{I}\models f_2$, then there exist two substitutions
			$\mu_1:\comp(f_1)\cap E\rightarrow C$ and $\mu_2:\comp(f_2)\cap E\rightarrow C$ such that $\mathfrak{I}\models f_1\mu_1$ and $\mathfrak{I}\models f_2\mu_2$.
			As the domains of the two substitutions are disjoint (by assumption), we can define the substitution $\mu:\comp(f)\cap E\rightarrow C$ as follows:
			\[\mu(v) = \begin{cases}
					\mu_1(v) & \text{ if } v\in \comp(f_1) \\
					\mu_2(v)& \, \text{else}
					\end{cases}\]
			Then $\mathfrak{I}\models f\mu$ and therefore $\mathfrak{I}\models f$.

	\end{enumerate}
	
	\end{proof}
If we recursively divide all pieces into sub-pieces, we get a maximal set $F=\{f_1, f_2, \ldots, f_n\}$  for each formula $f$ 
 such that $F\equiv \{f\}$ and for all
	 $1 \leq i, j \leq n$, 
	$\text{comp}(f_{i})\cap\text{comp}(f_{j})\cap E \neq \emptyset$ implies $i=j$.

Second, we replace all blank nodes occurring in rule bodies by \emph{fresh} universals. The rule 
\[\texttt{\{\_:x :likes :cake\}=>\{:cake :is :good\}.}\]
 becomes 
 \[\texttt{\{?y :likes :cake\}=>\{:cake :is :good\}.}\]
 Note that both rules have the same meaning, namely  \emph{``if someone likes cake, then cake is good''}.  We generalize that:
\begin{lemma}[Eliminating Existentials]\label{exel}\sloppy
	Let $f= \verb!{! e_1 \verb!}! \verb!=>! \verb!{! e_2 \verb!}!$ and $g=\verb!{! e'_1 \verb!}! \verb!=>! \verb!{! e_2 \verb!}!$ be  \nex implications such that $e'_1=e_1\sigma$ for some injective substitution $\sigma:\comp(e_1)\cap E\rightarrow U\setminus \comp(e_1)$
	of the existential variables of $e_1$ by universals.
	Then $f\equiv g$.
\end{lemma}
\begin{proof}
		We first note that $\comp(f)\cap E=\emptyset$ and $\comp(g)\cap E=\emptyset$ since both formulae are implications. 
	
	($\Rightarrow$)
	 We assume that $\mathfrak{M}\not\models g$ 
	 for some model $\mathfrak{M}$. That is, there exists a substitution $\nu: (\comp(e'_1)\cup\comp(e_2))\cap U\rightarrow C$ such that $\mathfrak{M}\models e'_1 \nu$ and 
	 $\mathfrak{M}\not\models e_2\nu$. We show that $\mathfrak{M}\models e_1\nu$: 
	As $((\comp(e_1)\cup\comp(e_2))\cap U)\subset ((\comp(e'_1)\cup\comp(e_2))\cap U)$, we know that $\comp(e_1\nu)\cap U =\emptyset$. With the substitution
	$\mu:= \nu \circ \sigma$
	for the existential variables in $e_1\nu$ we get $\mathfrak{M}\models (e_1 \nu) \sigma$ and thus $\mathfrak{M}\models (e_1 \nu)$, but as $\mathfrak{M}\not \models (e_2 \nu)$ we can conclude that $\mathfrak{M}\not \models f$.
	 
	($\Leftarrow$)
	 We assume that $\mathfrak{M}\not\models f$. That is, there exists a substitution $\nu: (\comp(e_1)\cup\comp(e_2))\cap U\rightarrow C$ such that $\mathfrak{M}\models e_1 \nu$ and 
	 $\mathfrak{M}\not\models e_2\nu$. 
	 As $\mathfrak{M}\models e_1 \nu$, there exists a substitution $\mu:\comp(e_1\nu)\cap E\rightarrow C$ such that $\mathfrak{M}\models (e_1 \nu)\mu$. With that we define a substitution $\nu':(\comp(e_1)\cup\comp(e_2))\cap U\rightarrow C$ as follows:
				 $\nu':U\rightarrow C$ as follows:
						\[\nu'(v) = \begin{cases}
					\mu(\sigma^{-1}(v)) & \text{ if } v\in range(\sigma) \\
					\nu(v)& \, \text{else}
				\end{cases}\]
	With that substitution we get $\mathfrak{M}\models e'_1\nu'$ but  $\mathfrak{M}\not\models e_2\nu'$ and thus $\mathfrak{M}\not\models g$.
\end{proof}
For a rule $f$  we call the formula $f'$ in which all existentials occurring in its body are replaced by universals following \Cref{exel} the \emph{normalized} version of the rule. We call an $\nthree{}^{\exists}$ formula $f$ \emph{normalized}, if all rules occurring in it as conjuncts are normalized. 
Combining the findings of the two previous lemmas, we introduce the \emph{Piece Normal Form}:
\begin{definition}[Piece Normal Form]\label{def:pnf}
	A finite set $\Phi = { f_1, f_2, \ldots, f_n }$ of $\nthree{}^\exists$ formulae is in \emph{piece normal form} (PNF) if all $f_i \in \Phi$ ($1 \leq i \leq n$) are normalized and $n \in \mathbb{N}$ is the maximal number such that for $1 \leq i,j \leq n$, $\comp(f_i) \cap \comp(f_j) \cap E \neq \emptyset$ implies $i = j$.
	If $f_i \in \Phi$ is a conjunction of atomic formulae, we call $f_i$ an \emph{atomic piece}. 
\end{definition}

We get the following result for \nex formulae:
\begin{theorem}\label{cor:pnf}
	For every well-formed \nex formula $f$, there exists a set $F =\{ f_{1}, f_{2}, \ldots, f_{k}\}$
	of \nex formulae such that $F\equiv \{f\}$ and $F$ is 
	in piece normal form. 
\end{theorem}
\begin{proof}
		The claim follows immediately from \Cref{lemma:pieces} and \Cref{exel}.
\end{proof}
Since the piece normal form $F$ of \nex formula $f$ is obtained by only replacing variables and separating conjuncts of $f$ into the set form, the overall size of $F$ is linear in $f$.

 \newcommand\termtf{\ensuremath{\mathbb{T}}\xspace}
\newcommand\domain{\ensuremath{\mathfrak{D}}\xspace}
\section{From N3 to Existential Rules}\label{exru}
Due to \Cref{cor:pnf}, we translate sets $F$ of \nex formulae in PNF (cf. \Cref{def:pnf}) to sets of existential rules $\transf(F)$ without loss of generality.
As a preliminary step, we introduce the language of existential rules formally.
Later on, we explain and formally define the translation function already sketched in \Cref{yea!}.
We close this section with a correctness argument, 
paving the way for existential rule reasoning for \nex formulae.
\subsection{Foundations of Existential Rule Reasoning}\label{sub:rule-prelims}
As for \nthree, we consider a first-order vocabulary, consisting of countably infinite mutually disjoint sets of constants (\constS), variables (\varS), and additionally so-called (labeled) nulls (\nullS)\footnote{We choose here different symbols to disambiguate between existential rules and \nthree, although vocabularies partially overlap.}. As already mentioned in \Cref{yea!}, we use the same set of constants as \nthree formulae, meaning $\constS = C$.
Furthermore, let $\predS$ be a (countably infinite) set of \emph{relation names}, where each $p\in\predS$ comes with an arity $\arity{p}\in\mathbb{N}$.
\predS is disjoint from the term sets \constS, \varS, and \nullS.
We reserve the ternary relation name $\tripl\in\predS$ for our encoding of \nthree triples.
If $p\in\predS$ and $t_{1},t_{2},\ldots,t_{\arity{p}}$ is a list of terms (i.e., each $t_{i}\in\constS\cup\nullS\cup\varS$), $p(t_{1},t_{2},\ldots,t_{\arity{p}})$ is called an \emph{atom}.
We often use $\vec t$ to summarize a term list like $t_{1},\ldots,t_{n}$ ($n\in\mathbb{N}$), and treat it as a set whenever order is irrelevant.
An atom $p(\vec t)$ is \emph{ground} if $\vec t \subseteq \constS$.
An \emph{instance} is a (possibly infinite) set $\Inst$ of variable-free atoms and a finite set of ground atoms $\Dnst$ is called a \emph{database}.

For a set of atoms $\Inst[A]$ and an instance $\Inst$, we call a function $h$ from the terms occurring in $\Inst[A]$ to the terms in $\Inst$ a \emph{homomorphism from $\Inst[A]$ to $\Inst$}, denoted by $h : \Inst[A]\to\Inst$, if (1) $h(c)=c$ for all $c\in\constS$ (occurring in \Inst[A]), and (2) $p(\vec t)\in\Inst[A]$ implies $p(h(\vec t))\in\Inst$.
If any homomorphism from \Inst[A] to \Inst exists, write $\Inst[A]\to\Inst$.
Please note that if $n$ is a null occurring in \Inst[A], then $h(n)$ may be a constant or null.

\newcommand\bdy[1]{\ensuremath{\textsf{body}(#1)}\xspace}
\newcommand\had[1]{\ensuremath{\textsf{head}(#1)}\xspace}
For an \emph{(existential) rule}
$\arule\colon \forall \vec{x}, \vec{y} .\ \varphi[\vec{x},\vec{y}] \rightarrow \exists \vec{z} .\ \psi[\vec{y},\vec{z}]$ (cf.\xspace \eqref{eq:exru-shape}),
rule body ($\bdy{\arule} := \varphi$) and head ($\had{\arule} := \psi$) will also be considered as sets of atoms for a more compact representation of the semantics.
The notation $\varphi[\vec x,\vec y]$ ($\psi[\vec y,\vec z]$, resp.) indicates that the only variables occurring in $\varphi$ ($\psi$, resp.) are $\vec x\cup \vec y$ ($\vec y\cup\vec z$, resp.).
A finite set of existential rules $\ruleS$ is called an \emph{(existential) rule program}.

Let $\arule$ be a rule and $\Inst$ an instance.
We call a homomorphism $h : \bdy{\arule} \to \Inst$ a \emph{match for $\arule$ in $\Inst$}.
Match $h$ is \emph{satisfied for \arule in \Inst} if there is an extension $h^{\star}$ of $h$ (i.e., $h\subseteq h^{\star}$) such that $h^{\star}(\had{\arule})\subseteq\Inst$.
If all matches of \arule are satisfied in \Inst, we say that \arule is satisfied in \Inst, denoted by $\Inst\models \arule$.
For a rule program $\ruleS$ and database \Dnst, instance \Inst is a \emph{model of \ruleS and \Dnst}, denoted by $\Inst \models \Sigma,\Dnst$, if $\Dnst\subseteq\Inst$ and $\Inst\models\arule$ for each $\arule\in\ruleS$.

Labeled nulls play the role of fresh constants without further specification, just like blank nodes in RDF or \nthree.
The chase is a family of algorithms that soundly produces models of rule programs by continuously applying rules for unsatisfied matches.
Rule heads are then instantiated and added to the instance.
Existentially quantified variables are thereby replaced by (globally) fresh nulls in order to facilitate arbitrary constant injections.
More formally, we call a sequence $\Dnst^0 \Dnst^1 \Dnst^2 \ldots$ a \emph{chase sequence of rule program $\ruleS$ and database \Dnst} if (1) $\Dnst^0 = \Dnst$ and (2) for $i>0$, $\Dnst^i$ is obtained from $\Dnst^{i-1}$ by applying a rule $\arule\in\ruleS$ for match $h$ in $\Dnst^{i-1}$ (i.e., $h : \bdy{r} \to \Dnst^{i-1}$ is an unsatisfied match and $\Dnst^i = \Dnst^{i-1} \cup \{ h^{\star}(\had{r}) \}$ for an extension $h^{\star}$ of $h$).
The \emph{chase of $\ruleS$ and \Dnst} is the limit of a chase sequence $\Dnst^0 \Dnst^1 \Dnst^2 \ldots$, i.e., $\bigcup_{i\geq 0} \Dnst^0$. 
Although chase sequences are not necessarily finite\footnote{This also means there is no guarantee of termination.}, the chase always is a (possibly infinite) model\footnote{Not just any model, but a universal model, which is a model that has a homomorphism to any other model of the database and rule program. Up to homomorphisms, universal models are unique, justifying the use of the article \emph{the} for \emph{the chase}.} \citep{DNR08:corechase}.
The described version of the chase is called \emph{standard chase} or \emph{restricted chase}.

We say that two rule programs $\ruleS_{1}$ and $\ruleS_{2}$ are \emph{equivalent}, denoted $\ruleS_{1} \Requiv \ruleS_{2}$, if for all instances $\Inst$, $\Inst\models\ruleS_{1}$ if and only if $\Inst\models\ruleS_{2}$.
Equivalences of existential rules have been extensively studied in the framework of data exchange \citep{FKNP2008:schemaMappingOptimization,PSS2011:relaxedSchemaMappingEquivalence}. Our equivalence is very strong and is called \emph{logical equivalence} in the data exchange literature.
For an alternative equivalence relation between rule programs, we could have equally considered equality of ground models (i.e., those models that are null-free).
Let us define this equivalence as follows: $\ruleS_{1} \Requiv_{g} \ruleS_{2}$ if for each ground instance $\Inst$, $\Inst\models\ruleS_{1}$ if and only if $\Inst\models\ruleS_{2}$.
The following lemma helps simplifying the proofs concerning the correctness of our transformation\footnote{We are going to use the terms \emph{translation} and \emph{transformation} synonymously throughout the rest of this section.} function later on.
\begin{lemma}\label{lemma:grounding}
  $\Requiv$ and $\Requiv_{g}$ coincide.
\end{lemma}
\begin{proof}
  Of course, $\Requiv\subseteq\Requiv_{g}$ holds since since the set of all ground models of a rule program is a subset of all models of that program.
  
  Towards showing $\Requiv_{g}\subseteq\Requiv$, assume rule programs $\ruleS_{1}$ and $\ruleS_{2}$ such that $\ruleS_{1}\Requiv_{g}\ruleS_{2}$, but $\ruleS_{1} \cancel\Requiv \ruleS_{2}$.
  Then there is a model $\Inst[M]$ of $\ruleS_{1}$, such that $\Inst[M]\cancel\models\ruleS_{2}$ (or vice versa), implying that for some rule $r\in\ruleS_{2}$ there is a match $h$ in $\Inst[M]$ but for no extension $h^{\star}$, we get $h^{\star}(\had{\arule})\subseteq\Inst[M]$.
  As $\ruleS_{1}\Requiv_{g} \ruleS_{2}$, $\Inst[M]$ cannot be a ground instance and, thus, contains at least one null.

  \textbf{Claim:} Because of $\Inst[M]$, there is a ground instance $\Inst[M]_{g}$, such that $\Inst[M]_{g}\models\ruleS_{1}$ and $\Inst[M]_{g}\cancel\models\ruleS_{2}$.
  But then $\Inst[M]_{g}$ constitutes a counterexample to the assumption that $\ruleS_{1}\Requiv_{g} \ruleS_{2}$. 
  Thus, the assumption $\ruleS_{1} \cancel\Requiv\ruleS_{2}$ would be disproven.

In order to show the claim, we construct $\Inst[M]_{g}$ from $\Inst[M]$ by replacing every null $n$ in $\Inst[M]$ by a (globally) fresh constant $c_{n}$.
  Since there might not be enough constants -- $\Inst[M]$ may already use all countably infinite constants $c\in\constS$ -- we take a little detour: although the set of constants in use might be infinite in $\Inst[M]$, the number of constants used inside the rule programs $\ruleS_{1}$ and $\ruleS_{2}$ is finite\footnote{Recall, a rule program is defined as a finite set of existential rules.}.
  Create instance $\Inst[M]''$ from $\Inst[M]$ by replacing all constants $c$ not part of $\ruleS_{1}$ or $\ruleS_{2}$ by fresh nulls $n_{c}$.
  Once again, $\Inst[M]$ may already use up all nulls $n\in\nullS$.
  So we have to take yet another detour from $\Inst[M]$ to $\Inst[M]'$ as follows: 
  
  Let $\gamma : \nullS\to\mathbb{N}$ be a (necessarily injective) enumeration of $\nullS$.
  Define $\eta : \constS\cup\nullS\to\constS\cup\nullS$ by (1) $\eta(c):=c$ for all $c\in\constS$ and (2) $\eta(n):= \eta^{-1}(2\cdot\eta(n))$.
  Then $\Inst[M]'$ is produced by applying $\eta$ to $\Inst[M]$.
  Note, for each number $k\in\mathbb{N}$, $\eta^{-1}(2k+1)$ is not a null in $\Inst[M]'$.
  Since $\eta$ is an isomorphism between $\Inst[M]$ and $\Inst[M]'$, we get that $\Inst[M]\models\ruleS$ if and only if $\Inst[M]'\models\ruleS$ for all rule programs $\ruleS$.
  Recall that isomorphic models preserve all first-order sentences (see, e.g., \citeauthor{Ebbinghaus1994}).
  Hence, $\Inst[M]'\models\ruleS_{1}$ and $\Inst[M]'\cancel\models\ruleS_{2}$.

  Next we construct $\Inst[M]''$ from $\Inst[M]'$ by function $\omega$ mapping the terms occurring in $\Inst[M]'$ to $\constS\cup\nullS$, such that (1) $\omega(c)=c$ if $c$ is a constant occurring in $\ruleS_{1}\cup\ruleS_{2}$, (2) $\omega(d)$ is a fresh null $n_{d}$ if $d$ is a constant not occurring in $\ruleS_{1}\cup\ruleS_{2}$, and (3) $\omega(n)=n$ otherwise.
  $\omega$ exists because the number of nulls in use by $\Inst[M]'$ is countably infinite.
  Note that $\omega$ is injective and $\omega(\Inst[M]')=\Inst[M]''$ uses only finitely many constants.
  Once again we show that $\Inst[M]'\models\ruleS$ if and only if $\Inst[M]''\models\ruleS$ for arbitrary rule programs $\ruleS$, implying that $\Inst[M]''\models\ruleS_{1}$ and $\Inst[M]''\cancel\models\ruleS_{2}$:

  Let $r\in\ruleS$ with match $h$ in $\Inst[M]'$.
  If $h$ is satisfied in $\Inst[M]'$, then there is an extension $h^{\star}$, such that $h^{\star}(\had{r})\subseteq\Inst[M]'$.
  By definition of $\omega$ and, thus, the construction of $\Inst[M]''$, $\omega \circ h$ is a match for $r$ in $\Inst[M]''$ and $\omega\circ h^{\star}$ its extension with $\omega\circ h^{\star}(\had{r})\subseteq\Inst[M]''$.
  The converse direction uses the the same argumentation, now from $\Inst[M]''$ to $\Inst[M]'$, using the fact that $\omega$ is injective.

  From $\Inst[M]''$ we can finally construct ground instance $\Inst[M]_{g}$ by $\nu$ mapping all (finitely many) constants $c$ in $\Inst[M]''$ to themselves and every null $n$ in $\Inst[M]''$ to a fresh constant $c_{n}$.
  It holds that $\Inst[M]''\models\ruleS$ if and only if $\nu(\Inst[M]'')=\Inst[M]_{g} \models \ruleS$ (for all rule programs $\ruleS$) by a similar argumentation as given in the step from $\Inst[M]'$ to $\Inst[M]''$ above.
  Thus, $\Inst[M]_{g}\models\ruleS_{1}$ and $\Inst[M]_{g}\cancel\models\ruleS_{2}$, which completes proof.
\end{proof}

\subsection{The Translation Function from N3 to Existential Rules}
The translation function $\transf$ maps sets $F = \{ f_{1}, \ldots, f_{k} \}$ of \nex formulae in PNF to existential rule programs $\ruleS$.
Before going into the intricates of $\transf$ for every type of piece, consider the auxiliary function $\termtf : C\cup E\cup U \to \constS\cup\varS$ mapping \nthree terms to terms in our rule language (cf. previous subsection):\\[.3em]
\centerline{$\termtf(t) := \left\{
  \begin{array}{ccl}
    \uvar{x} && \text{if $t=\texttt{?x}\in U$}  \\
    \evar{y} && \text{if $t=\texttt{\_:y}\in E$} \\
    t && \text{if $t\in C$,}
  \end{array}
\right.$}\\[.3em]
where $\uvar{x},\evar{y}\in\varS$ and $t\in\constS$ (recall that $C=\constS$).
While variables in \nthree belong to either $E$ or $U$, this separation is lost under function $\termtf$.
For enhancing readability of subsequent examples, the identity of variables preserves this information by using superscripts $\exists$ and $\forall$.
Function \termtf naturally extends to triples $g = \texttt{$t^{1}$ $t^{2}$ $t^{3}$}$: $\termtf(g):=\tripl(\termtf(t^{1}), \termtf(t^{2}), \termtf(t^{3}))$.
We provide the translation for every piece $f_{i}\in F$ ($1\leq i\leq k$) and later collect the full translation of $F$ as the union of its translated pieces.

\paragraph{Translating Atomic Pieces.}
If $f_{i}$ is an atomic piece, $f_{i} = \texttt{$g_{1}$ $g_{2}$ $\ldots$ $g_{l}$}$ for some $l\geq 1$ and each $g_{j}$ ($1\leq j\leq l$) is an atomic formula.
The translation of $f_{i}$ is the singleton set $\transf(f_{i})=\{ \rightarrow \exists \vec z .\ \termtf(g_{1})\wedge \termtf(g_{2})\wedge \ldots\wedge \termtf(g_{l}) \}$, where $\vec z$ is the list of translated existential variables (via \termtf) from existentials occurring in $f_i$.
For example, the formula in \eqref{eq:2bl} constitutes a single piece $f_{\eqref{eq:2bl}}$ which translates to $$\transf(f_{\eqref{eq:2bl}}) = \left\{ \rightarrow \exists \evar{\ntt{y}} .\ \tripl(\ntt{:lucy},\ntt{:knows},\evar{\ntt{y}}) \wedge \tripl(\evar{\ntt{y}},\ntt{:likes},\ntt{:cake}) \right\}\text.$$\paragraph{Translating Rules.}
For rule-shaped pieces $f_{i} = \texttt{\{$e_1$\}=>\{$e_2$\}}$, we also obtain a single existential rule.
Recall that our PNF ensures all variables in $e_{1}$ to be universals and all universal variables in $e_{2}$ to also occur in $e_{1}$.
If $e_{1} = \texttt{$g_{1}^{1}$ $g_{1}^{2}$ $\cdots$ $g_{1}^{m}$}$ and $e_{2} = \texttt{$g_{2}^{1}$ $g_{2}^{2}$ $\cdots$ $g_{2}^{n}$}$, $\transf(f_{i}):=\{ \forall \vec x .\ \bigwedge_{j=1}^{m} \termtf(g_{1}^{j}) \rightarrow \exists \vec z .\ \bigwedge_{j=1}^{n} \termtf(g_{2}^{j}) \}$ where $\vec x$ and $\vec z$ are the lists of translated universals and existentials, respectively.
Applying \transf to the \nthree formula in \eqref{eq:exq-n3}, which is a piece according to \Cref{def:pnf}, we obtain
$$\transf(f_{\eqref{eq:exq-n3}}) = \left\{ \forall \uvar{\ntt{x}}.\ \tripl(\uvar{\ntt{x}}, \ntt{:knows}, \ntt{:tom}) \rightarrow \exists \evar{\ntt{y}} .\ \tripl(\uvar{\ntt{x}}, \ntt{:knows}, \evar{\ntt{y}}) \wedge \tripl(\evar{\ntt{y}}, \ntt{:name}, \ntt{"Tom"}) \right\}\text,$$
which is the same rule as given in \eqref{eq:exq-rl} up to a renaming of (bound) variables.

\paragraph{Translating the PNF.}
For a set $F = \{f_{1}, f_{2}, \ldots, f_{k}\}$ of \nex formulae in PNF, $\transf(F)$ is the union of all translated constituents (i.e., $\transf(F) := \bigcup_{i=1}^{k} \transf(f_{i})$).
Please note that \transf does not exceed a polynomial overhead in its input size.

\subsection{Correctness of the Translation}

Let $F$ be a set of \nex formulae in PNF.
Its translation $\transf(F)$ provides the following soundness guarantee: For every ground model $\Inst[M]$ of $\transf(F)$, there is an interpretation of $F$ that is itself a model.

\begin{theorem}\label{lemma:negotiation}
  Let $F$ be a set of \nex formulae in PNF and $\Inst[M]$ a ground instance.
  Define \emph{the canonical interpretation of $\Inst[M]$} by $\mathfrak{I}(\Inst[M]) = (C, \mathfrak{a}, \mathfrak{p})$ such that
  \begin{itemize}
\item $\mathfrak{a}(t) := t$ for all $t\in C$ and
\item $\mathfrak{p}(p) := \{ (s,o) \mid \tripl(s,p,o)\in\Inst[M] \}$ for all $p\in C$.
  \end{itemize}
  $\Inst[M]$ is a model of $\transf(F)$ if and only if $\mathfrak{I}(\Inst[M])$ is a model of $F$.
\end{theorem}

\begin{proof}
By induction on the number $k$ of pieces in $F = \{ f_{1}, f_{2}, \ldots, f_{k} \}$:
  \begin{description}
    \item[Base:] For $k=1$, $F = \{ f \}$ and $f$ is either (a) an atomic piece or (b) a rule, being the cases we need to distinguish.
      \begin{enumerate}[label=(\alph*)]
        \item $\transf(F)=\transf(f)=\{ \rightarrow \exists \vec z .\ \bigwedge_{i=1}^{n} \tripl(s_{i}, p_{i}, o_{i}) \}$.
          Every model of $\transf(F)$ satisfies its single rule, meaning that if $\Inst[M]$ is a model, there is a homomorphism $h^{\star}$ from $\mathcal{A} = \{ \tripl(s_{i},p_{i},o_{i}) \mid 1\leq i\leq n \}$ to $\Inst[M]$.
          From $\Inst[M]$ we get $\mathfrak{I}(\Inst[M])=(C,\mathfrak{a},\mathfrak{p})$ with $(s_{i},o_{i})\in\mathfrak{p}(p_{i})$ for all $i\in\{ 1,\ldots,n \}$. We need to show that $\mathfrak{I}(\Inst[M])$ is a model for $F$ (i.e., for $f$).

          If $f$ contains existentials (i.e., $\text{comp}(f)\cap E = W$ is nonempty), we need to find a substitution $\mu : W \to C$ such that $\mathfrak{I}(\Inst[M]) \models f\mu$.
          Define $\mu : W \to C$ alongside $h^{\star}$: $\mu(\texttt{\_:y})=h^{\star}(\evar{y})$ for each $\ntt{\_:y}\in W$.
For each atomic formula $g_{j} = \texttt{$s_{j}$ $p_{j}$ $o_{j}$}$ of $f$, we get $\mathfrak{I}(\Inst[M]) \models g_{j}\mu$ since $\tripl(h^{\star}(s_{j}),h^{\star}(p_{j}),h^{\star}(o_{j}))\in\Inst[M]$ implies $(h^{\star}(s_{j}),h^{\star}(o_{j}))\in\mathfrak{p}(h^{\star}(p_{j}))$ and, thus, $(\mathfrak{a}(s_{j}\mu),\mathfrak{a}(o_{j}\mu))\in\mathfrak{p}(\mathfrak{a}(p_{j}\mu))$.
          This argument holds for every atomic formula $g_{j}$ of $f$, implying $\mathfrak{I}(\Inst[M])\models F$.
          The converse direction uses the same argumentation backwards, constructing $h^{\star}$ from $\mu$.

          If $f$ does not contain any existentials, then $f$ is itself an atomic formula and the result follows as the special case $n=1$.
\item If $F = \{ f \}$ and $f= \texttt{\{$e_1$\}=>\{$e_2$\}}$, then $\transf(F)=\{ \forall \vec x .\ \varphi \rightarrow \exists \vec z .\ \psi \}$ where $\varphi$ and $\psi$ are translated conjunctions from $e_{1}$ and $e_{2}$.
        
          Let $\mathfrak{I}(\Inst[M])$ be a model of $F$.
          To show that $\Inst[M]$ is a model of $\transf(F)$, it suffices to prove, for each match $h$ of the rule, the existence of an extension $h^{\star}$ (of $h$), such that $h^{\star}(\psi)\subseteq\Inst[M]$. 

          Let $h$ be a match for the body of the rule and the body of the rule is a conjunction of atoms.
          Define $\sigma$ with $\sigma(\ntt{?x}):= h(\uvar{x})$ for each universal variable in $e_{1}$.
          $\sigma$ is a substitution such that $\mathfrak{I}(\Inst[M])\models e_{1}\sigma$.
          In order to prove this claim, let $\texttt{$s$ $p$ $o$}$ be a triple in $e_{1}$.
          Hence, $\tripl(s, p, o)\in\varphi$ and, by the choice of $h$, $\tripl(h(s), h(p), h(o))\in \Inst[M]$.
          This implies that $(h(s), h(o))\in \mathfrak{p}(h(p))$, which also implies $(s\sigma, o\sigma)\in\mathfrak{p}(p\sigma)$.
          As this argument holds for all triples in $e_{1}$, the claim follows.
          Please note that, as in case (a), this reasoning can be converted to construct a match $h$ from a substitution $\sigma$.

          Since $\mathfrak{I}(\Inst[M])$ is a model of $f$, there is a substitution $\mu : \text{comp}(e_{2})\cap E \to C$, such that $\mathfrak{I}(\Inst[M])\models e_{2}\sigma\mu$.
          Define $h^{\star} := h \cup \{ w\mapsto \mu(w) \mid w\in\text{comp}(e_{2})\cap E \}$.
          It holds that $h^{\star}$ satisfies match $h$ since for each atomic formula $\texttt{$s_{i}$ $p_{i}$ $o_{i}$}$ of $e_{2}$, we get $\mathfrak{a}(\mu(\sigma(s_{i})),\mu(\sigma(o_{i})))\in\mathfrak{p}(\mathfrak{a}(\mu(\sigma(p_{i}))))$ implying $\tripl(\mu(\sigma(s_{i})), \mu(\sigma(p_{i}),    \mu(\sigma(o_{i}))))\in\Inst[M]$ and $h^{\star}(\termtf(x))=\mu(\sigma(x))$ ($x\in\{ s_{i}, p_{i}, o_{i} \}$) providing a match for $\tripl(\termtf(s_{i}), \termtf(p_{i}), \termtf(o_{i}))$ (part of the head $\psi$).
As this argument holds for all atomic formulae of $e_{2}$, match $h$ is satisfied via $h^{\star}$.
          As before, the construction can be inverted, obtaining $\mu$ from $h^{\star}$ and $\sigma$ from $h$, which completes the proof for this case.
      \end{enumerate}

\item[Step:] Let $F = \{ f_{1}, f_{2}, \ldots, f_{k}, f_{k+1} \}$ be a set of \nex formulae in PNF.
By induction hypothesis, $\Inst[M]$ is a model of $\transf(\{ f_{1}, f_{2}, \ldots, f_{k} \})$ if and only if $\mathfrak{I}(\Inst[M])$ is a model of $\{ f_{1}, f_{2}, \cdots, f_{k} \}$.
          Also by induction hypothesis, $\Inst[M]$ is a model of $\transf(\{ f_{k+1} \})$ iff $\mathfrak{I}(\Inst[M])$ is a model of $\{ f_{k+1} \}$.
          Thus, $\Inst[M]$ is a model of $\transf(F)$ if and only if it is a model of $\transf(\{ f_{1} f_{2} \cdots f_{k} \})$ and of $\transf(\{ f_{k+1} \})$ if and only if $\mathfrak{I}(\Inst[M])$ is a model of $\{ f_{1} f_{2} \cdots f_{k} \}$ and of $\{ f_{k+1} \}$ if and only if $\mathfrak{I}(\Inst[M])$ is a model of $F$.
\end{description}
\end{proof}
Consequently, the only (ground) models $\transf(F)$ has are models of the original set of \nex formulae $F$.
To complete the correctness argument, $\transf(F)$ can have only those models relating to the ones of $F$, which is also true by the following theorem.

\begin{theorem}\label{thm:full-abstraction}
Let   $\mathfrak{M}$ be an \nthree interpretation, then
there exists an instance $\Inst[M]$ such that
for each  set $F$ of \nex formulae in PNF 
\[\mathfrak{M}\models F \text{ if and only if } \mathfrak{I}(\Inst[M])\models F\] (cf.\xspace \Cref{lemma:negotiation}). 
\end{theorem}
\begin{proof}

For $\mathfrak{M}=(\domain,\mathfrak{a},\mathfrak{p})$, define $\mathfrak{M}_g=(C,\mathfrak{b},\mathfrak{q})$ such that (a) $\mathfrak{b}$ is the identity on $C$ (i.e., $\mathfrak{b}(c)=c$ for all $c \in C$) and (b) $\mathfrak{q}(p) := \{ (s,o) \mid (\mathfrak{a}(s),\mathfrak{a}(o))\in\mathfrak{p}(\mathfrak{a}(p)) \}$ for all $p\in C$.
          Based on $\mathfrak{M}_g$, we can define instance $\Inst[M] := \{ \tripl(s,p,o) \mid (s,o)\in\mathfrak{q}(p) \}$.

          Since $\mathfrak{I}(\Inst[M]) = \mathfrak{M}_g$, it remains to be shown that $\mathfrak{M}\models F$ if and only if $\mathfrak{M}_g \models F$.
          We proceed by induction on (the number of pieces) $|F|=k$.
          \begin{description}
            \item[Base:] If $k=1$, then $F = \{ f \}$ and two cases arise: (a) $f$ is an atomic piece $\texttt{$g_1$ $\cdots$ $g_l$}$ (for some $l \geq 1$) and (b) $f$ is an \nthree rule $\texttt{\{$e_1$\}=>\{$e_2$\}}$.
              \begin{enumerate}[label=(\alph*)]
                \item $\mathfrak{M}\models f$ if and only if $\mathfrak{M}\models f\mu$ for some $\mu : \text{comp}(f)\cap E \to C$ if and only if for each atomic formula $g_i = \texttt{$s_i$ $p_i$ $o_i$}$ in $f$, $(\mathfrak{a}(\mu(s_i)),\mathfrak{a}(\mu(o_i)))\in\mathfrak{p}(\mathfrak{a}(\mu(p_i)))$ (by the semantics of \nex) if and only if $(\mathfrak{b}(\mu(s_i)),\mathfrak{b}(\mu(o_i)))\in\mathfrak{q}(\mathfrak{b}(\mu(p_i)))$ (by construction of $\mathfrak{M}_g$) if and only if $\mathfrak{M}_g\models f\mu$ if and only if $\mathfrak{M}_g\models f$ (by the semantics of \nex).
                \item $\mathfrak{M}\models f$
if and only if, for each substitution $\sigma : U \to C$ with $\mathfrak{M}\models e_{1}\sigma$, there is a substitution $\mu : \text{comp}(e_2) \cap E \to C$ such that $\mathfrak{M}\models e_{2}\sigma\mu$.
                   
                  For respective substitutions $\sigma : U \to C$ and $\mu : \text{comp}(e_{2})\cap E \to C$,
                  $\mathfrak{M}\models e_{1}\sigma$ if and only if $(\mathfrak{a}(\sigma(s)),\mathfrak{a}(\sigma(o)))\in\mathfrak{p}(\mathfrak{a}(\sigma(p)))$ for each atomic formula $\texttt{$s$ $p$ $o$}$ in $e_{1}$ (by the semantics of \nex) if and only if $(\mathfrak{b}(\sigma(s)),\mathfrak{b}(\sigma(o)))\in\mathfrak{q}(\mathfrak{b}(\sigma(p)))$ for each atomic formula $\texttt{$s$ $p$ $o$}$ of $e_{1}$ if and only if $\mathfrak{M}_g\models e_{1}\sigma$.
                  
                  The same argument can be used to argue for $\mathfrak{M}\models e_{2}\sigma\mu$ if and only if $\mathfrak{M}_g\models e_{2}\sigma\mu$.
                  Thus, for each $\sigma : U \to C$ for which $\mathfrak{M}\models e_{1}\sigma$ there is a substitution $\mu : \text{comp}(e_{2})\cap E \to C$ such that $\mathfrak{M}\models e_{2}\sigma\mu$ and we obtain $\mathfrak{M}_g\models e_{1}\sigma$ and $\mathfrak{M}_g\models e_{2}\sigma\mu$, and vice versa.
              \end{enumerate}
            \item[Step:] For $F=\{ f_{1}, \ldots, f_{k}, f_{k+1} \}$, the induction hypothesis applies to $F'=\{ f_{1}, \ldots, f_{k} \}$ and $F'' = \{ f_{k+1} \}$: \begin{align*}
\mathfrak{M}\models F\quad & \text{iff}\quad  \mathfrak{M}\models F' \text{ and } \mathfrak{M}\models F'' &\text{(by \Cref{lemma:pieces})}\\
&\text{iff}\quad  \mathfrak{M}_g\models F' \text{ and } \mathfrak{M}_g\models F'' & \text{(by induction hypothesis)}\\
&  \text{iff}\quad  \mathfrak{M}_g\models  F & \text{ (by \Cref{lemma:pieces})}
\end{align*}
\end{description}

\end{proof}
In the conference version of this paper, we have shown the following result to justify correctness of \transf, relating equivalent \nthree formulae to their translations.
\begin{corollary}[Theorem 2 \citep{rr23n3}]\label{oldthm:full-abstraction}
  For sets of \nex formulae $F$ and $G$ in PNF, $F \Ntequiv G$ if and only if $\transf(F) \Requiv \transf(G)$.
\end{corollary}
This kind of correctness argument has been coined to the term of \emph{full abstraction} \citep{GN2016:fullAbstractionHistoryMyths}.
\begin{proof}
  Let $F$ and $G$ be sets of \nex formulae in PNF.
  The proof disects the statement in its two logical parts:
\begin{description}
  \item[Soundness ($\Rightarrow$)] Assume $F \Ntequiv G$.
  We need to show that $\transf(F) \Requiv \transf(G)$.
  Let $\Inst[M]$ be a ground model of $\transf(F)$. 
  Then by \Cref{lemma:negotiation}, $\mathfrak{I}(\Inst[M])$ is a model of $F$. 
  By assumption ($F \Ntequiv G$), $\mathfrak{I}(\Inst[M])$ is a model of $G$ and, again by an application of \Cref{lemma:negotiation}, we get that $\Inst[M]$ must be a model of $\transf(G)$.
  Hence, $\transf(F) \Requiv_g \transf(G)$ which implies $\transf(F) \Requiv \transf(G)$ by \Cref{lemma:grounding}.
  The converse direction, starting from ground models of $\transf(G)$, uses the inverse argumentation. 
  \item[Completeness ($\Leftarrow$)] We prove the contrapositive: $F \cancel{\Ntequiv} G$ implies $\transf(F) \cancel{\Requiv} \transf(G)$.
  Assume $F \cancel{\Ntequiv} G$.
  Then there is a model $\mathfrak{M}$ such that $\mathfrak{M} \models F$ and $\mathfrak{M} \cancel{\models} G$, or vice versa.
  Since the cases are symmetric, without loss of generality, assume $\mathfrak{M} \models F$ and $\mathfrak{M} \cancel{\models} G$ and suppose, towards a contradiction, that $\mathbf{(\star)}$ $\transf(F) \Requiv \transf(G)$.
  By \Cref{thm:full-abstraction}, there is an instance $\Inst[M]$ such that $\mathfrak{M} \models H$ if and only if $\mathfrak{I}(\Inst[M]) \models H$ for arbitrary sets $H$ of \nex formulae in PNF.
  Thus, $\mathfrak{I}(\Inst[M]) \models F$ and $\mathbf{(\star\star)}$ $\mathfrak{I}(\Inst[M]) \cancel{\models} G$.
  By \Cref{lemma:negotiation}, we get that $\Inst[M] \models \transf(F)$ and, by assumption $\mathbf{(\star)}$, $\Inst[M] \models \transf(G)$.
  But then \Cref{lemma:negotiation} allows for the conclusion that $\mathfrak{I}(\Inst[M]) \models G$, contradicting $\mathbf{(\star\star)}$.
  Therefore, the assumption $\mathbf{(\star)}$ is false, meaning $\transf(F) \cancel{\Requiv}\transf(G)$.
\end{description}
\end{proof}
In the conference version of this paper, \Cref{oldthm:full-abstraction} originated from the attempt to evade trivial and/or \emph{unexpectedly simple yet undesirable} transformations, which our transformation function \transf does not belong to because it possesses even stronger guarantees as proven in \Cref{lemma:negotiation,thm:full-abstraction}.
For instance, consider a translation $\transf_0$ mapping all sets of \nex formulae in PNF to some fixed set of rules $\Sigma_0$ (e.g., $\Sigma_0 = \emptyset$).
This translation is surely \emph{sound} since the image of $\transf_0$ consists of the single set of rules $\Sigma_0$, being trivially equivalent to itself.
On the other hand, requiring \emph{completeness} rules out $\transf_0$ as a \emph{good} transformation, since also non-equivalent \nthree formulae map to the same (i.e., equivalent) rule program.

While $\transf_0$ can be ruled out as trivially incomplete by \Cref{oldthm:full-abstraction}, undesirable transformations are still in reach:
Consider an arbitrary enumeration of equivalence classes of sets of \nex formulae in PNF $\mathfrak{F}_1, \mathfrak{F}_2, \mathfrak{F}_3, \ldots$ and for each of these classes a first (e.g., the lexicographically smallest) element $\mathbf{1^{st}}(\mathfrak{F}_i)$.
Define $\transf_1(F) := \transf(\mathbf{1^{st}}(\mathfrak{F}_{i+1}))$ if $F \in \mathfrak{F}_i$.
$\transf_1$ is \emph{sound} and \emph{complete} in the sense of \Cref{oldthm:full-abstraction}, but fails in having stronger guarantees on the relationship between the different semantic worlds.
$\transf_1(F)$ may end up with a rule program that speaks about a completely different vocabulary -- in consequence, completely different subjects -- than $F$ does.
Even worse, the \emph{correctness} of $\transf_1$ does not even require the intermediate transformation \transf to be correct in any way:
different equivalence classes must just be associated with (semantically) different rule programs.

$\transf_0$ is \emph{incomplete} with respect to \Cref{oldthm:full-abstraction}.
$\transf_1$ on the other hand satisfies \Cref{oldthm:full-abstraction} but it does not share the same strong characteristics of \transf proven in \Cref{lemma:negotiation,thm:full-abstraction}:
Models of translated formulas cannot easily be converted into models of the original \nthree formula, making such transformations hard to use in contexts in which we want to employ existential rule reasoning as alternatives to existing \nthree reasoners.
In retrospect, the connection between \nthree models and models of existential rules through \transf turns out to be much deeper than captured by \Cref{oldthm:full-abstraction} alone.
This depth has been exploited within the proof of \Cref{oldthm:full-abstraction} that used to be available in our technical appendix \citep{arxiv2023report} only.
Nevertheless, most parts of the original proofs are kept and allow for the stronger statements we now describe by \Cref{lemma:negotiation,thm:full-abstraction}.
The new and stronger formulation is meaningful to \emph{reasoning} since all reasoning results can be translated back-and-forth through \transf.
It is the combination of \Cref{lemma:negotiation,thm:full-abstraction,oldthm:full-abstraction} that makes our function \transf a useful contribution.
Note that similar (and further) issues regarding \emph{full abstraction} have been uncovered in other communities before \citep{GN2016:fullAbstractionHistoryMyths,Parrow2016:fullAbstractionForFree}.

Apart from correctness of \transf and the close connection between \nthree models and models of transformed \nex formulae, we have no further guarantees.
As \nex reasoning does not necessarily stop, there is no need for termination of the chase over translated rule programs.
We expect that the similarity between the identified \nthree fragment of \nex and existential rules allows for the adoption of sufficient conditions for finite models, for instance, by means of acyclicity (see \citeauthor{CG+13:acyclicity} for a survey).

 \section{Reasoning with Lists}\label{sec:lists}
So far, we discussed $\nthree^{\exists}$ as a fragment of \nthree which can directly be mapped to existential rules. 
In this section, we detail how  $\nthree^{\exists}$ and our translation to existential rules can be extended towards supporting lists.
Lists is a very important concept in \nthree. 
We first explain them in more detail and provide  their semantics.
Then we explain how lists and list functions can be covered by existential rules. 
We finish the section by discussing different ways to implement list functions in \nthree.

\subsection{\nthree Lists}\label{sub:n3lists}
Before introducing them formally, we explain the role of lists in Notation3 Logic by examples. 
\nthree is based on RDF, but, in contrast to RDF, N3 treats lists as first-class citizens.  To  illustrate this, we take a closer look 
at the following  triple containing a list:

\begin{equation}
	\texttt{:lucy :likes (:cake :chocolate :tea).} \label{list}
\end{equation}

Stating that lucy likes cake, chocolate and tea.  If we understand the above as an example of RDF-turtle \citep{Turtle08}, the list-notation \texttt{( )}
is syntactic sugar for:

\begin{align}
&	\texttt{:lucy :likes \_:l1. }\label{listrdf} \\ & \texttt{\_:l1 rdf:first :cake; rdf:rest \_:l2.}\notag  \\ &\texttt{\_:l2 rdf:first :chocolate; rdf:rest \_l3.}\notag  \\&\texttt{\_:l3 rdf:first :tea; rdf:rest rdf:nil.} \notag
\end{align}

According to RDF semantics the predicates \texttt{rdf:first} and \texttt{rdf:rest}  are properties whose domain is the class of lists,  for \texttt{rdf:rest} the range is the class of lists and  \texttt{rdf:nil} is itself a list.  Their meaning is not specified any further.

In \nthree, the list in \eqref{list} itself is understood as a resource and not just as syntactic sugar for \eqref{listrdf}. The predicates \texttt{rdf:first} and \texttt{rdf:rest} have a more specific meaning: they stand for the relation between a list and its first element, a list and its rest list, that is the list, we retrive if we remove the first element, respectively. The rule
\begin{equation}
	\texttt{\{ (:a :b :c) rdf:first ?x; rdf:rest ?y\}=>\{?x :and ?y\}.}
\end{equation}
for example, yields
\begin{equation}
	\texttt{ :a :and (:b :c).}
\end{equation}
The constant \texttt{rdf:nil} stands for the empty list and can also be written as \texttt{( )}. 

If we define the semantics in a naive way, \nthree's view of lists is not fully compatible with the (syntactic sugar) view of RDF. Suppose, we have a new triple stating the food preferences of Tom (which coincide with Lucy's preferences):
\begin{equation}
	\texttt{:tom :likes (:cake :chocolate :tea).} \label{list2}
\end{equation}
If we apply the \nthree rule
\begin{equation}\label{prule}
	\texttt{\{?x :likes ?z. ?y :likes ?z\}=>\{?x :sharesPreferencesWith ?y\}.}
\end{equation}
on triple \eqref{list2} and \eqref{list}, we retrieve\footnote{Of course, we retrieve more, namely, that Tom shares preferences with Lucy and that both share preferences with themselves.} that
\begin{equation}\label{share}
	\texttt{:lucy :sharesPreferencesWith :tom.}
\end{equation}
Now, we replace \eqref{list2} by the first-rest combination it stands for, namely
\begin{align}
	&	\texttt{:ben :likes \_:k1. }\label{listben} \\ & \texttt{\_:k1 rdf:first :cake; rdf:rest \_:k2.}\notag  \\ &\texttt{\_:k2 rdf:first :chocolate; rdf:rest \_:k3.}\notag  \\&\texttt{\_:k3 rdf:first :tea; rdf:rest rdf:nil.} \notag
\end{align}
If we again apply rule \eqref{prule}, but this time on the list representations \eqref{listrdf} and \eqref{listben}, it is not evident that we get triple \eqref{share} as a result. The lists are represented by blank nodes \texttt{\_:l1} and \texttt{\_:k1}, and it is not immediately clear that these refer to the same list.
The original informal \nthree specification overcomes the problems caused by the different representations by providing
the following three axioms \citep{BernersLeeCKH08:N3Logic:tplp,BernersLeeC11:n3team}:
\begin{description}
	\item[Existence of Lists]  All lists exist.  That is, the triple
   \texttt{[rdf:first :a; rdf:rest rdf:nil].} does not carry any new information.
	\item[Uniqueness of Lists] Two lists having the same \texttt{rdf:first}-element and also the same \texttt{rdf:rest}-element are equal. If we add the notion of equality\footnote{Note that this equality is not that same kind of equality that the \nthree predicate \texttt{log:equalTo} provides. The latter is on syntax and not on the semantic level.} (\texttt{=}):
	\texttt{\{?L1 rdf:first ?X; rdf:rest ?R.
			?L12 rdf:first ?X; rdf:rest ?R.\} => \{?L1 = ?L2\}.}
	\item[Functionality] The predicates \texttt{rdf:first} and \texttt{rdf:rest} are functional properties. If we again add equality (\texttt{=}):\\
	\texttt{\{?S rdf:first ?O1, ?O2.\}=>\{?O1 = ?O2\}.}\\
	\texttt{\{?S rdf:rest ?O1, ?O2.\}=>\{?O1 = ?O2\}.}
\end{description}
The first axiom guarantees that there is no new informaion added when translating from the native list notation \eqref{list} to the first-rest noation \eqref{listrdf}.  The second and the third are important for the other direction, and, in a modified version, also for the purposes of 
our research which is to express \nthree lists and list predicates by means of existential rules. We will come back to that in \Cref{exlist}. 

Before introducing the non-basic list predicates, we provide the syntax and semantics of the extension of $\nthree{}^{\exists}$ with basic lists. We start with the syntax and extend the grammar provided in \Cref{N3S} as follows:
\begin{itemize}
	\item the set $\texttt{t}$ of term additionally contains the  empty list \texttt{()} and the concept \texttt{(l)} of list terms, with 
	   \begin{align*}
	   	\texttt{l ::= }&\\
	   	& \texttt{t}\\
	   	& \texttt{l t}	   	
	   	\end{align*}
		\item the set $\texttt{n}$ of N3 terms additionally contains the concept \texttt{(k)} of N3 list terms, with 
	\begin{align*}
		\texttt{k ::= }&\\
		& \texttt{n}\\
		& \texttt{k n}	   	
	\end{align*}
\end{itemize}

We further need to extend the \emph{application} of a substitution introduced in \Cref{n3semantics} by
$\texttt{(t}_1 \ldots \texttt{t}_n\texttt{)}\sigma = \texttt{(t}_1\sigma \ldots \texttt{t}_n\sigma\texttt{)}$ if $x= \texttt{(t}_1 \ldots \texttt{t}_n\texttt{)}$ is a list,
and the object function $\mathfrak{a}$ of N3 interpretations $\mathfrak{I}=(\mathfrak{D}, \mathfrak{a}, \mathfrak{p})$ as follows:
 If $\texttt{t}=\texttt{(t}_1 \ldots \texttt{t}_n\texttt{)}$  then $\mathfrak{a}(\texttt{t})=(\mathfrak{a} (\texttt{t}_1) \ldots \mathfrak{a}(\texttt{t}_n))$.  If $\texttt{t}=\texttt{()}$ then $\mathfrak{a}(\texttt{t})=()$. 
 
Note, that with our extension the domain $\mathfrak{D}$ of a model for a graph containing a list term also needs to contain a list of domain elements. However, the number of lists necessarily contained in $\mathfrak{D}$  is determined by the number of lists which can be produced using the alphabet. It is countable and does not depend on $\mathfrak{D}$ itself.  If $\mathfrak{D}$ contains all lists which can be constructed using the interpretations of the \nthree terms, then axiom 1 (existence of lists) is fulfilled.

We finish the definition of the semantics of $\nthree{}^{\exists}$ with basic lists as follows:

Given an \nthree alphabet which contains the list constants  \texttt{rdf:first} and \texttt{rdf:rest}, and an \nthree Interpretation $\mathfrak{I}=(\mathfrak{D}, \mathfrak{a}, \mathfrak{p})$.
We say that $\mathfrak{I}$ is a model according to the simple list semantics of a formula $\phi$, written as $\mathfrak{I}\models_{sl} \phi$ iff  $\mathfrak{I}\models\phi$ and for triples containing  \texttt{rdf:first} or \texttt{rdf:rest} in predicate position:
\begin{itemize}
	\item  $\mathfrak{I}\models_{sl}  s\ \texttt{rdf:first}\  o.$ iff $\mathfrak{a}(s)=(s_1 \ldots s_n)$ and $\mathfrak{a}(o)=s_1$
	\item  $\mathfrak{I}\models_{sl} s\ \texttt{rdf:rest}\ o.$ iff $\mathfrak{a}(s)=(s_1 \, s_2\ldots s_n)$ and $\mathfrak{a}(o)=(  s_2\ldots s_n)$
\end{itemize}
Note that with this definition, we also fulfill the two missing axioms stated above. The syntactic list structure maps to a list structure in the domain of discourse. This domain list can only have one first element and only one rest list, and it is fully  determined  by these two parts.

In addition to \texttt{rdf:first} and \texttt{rdf:rest}, \nthree contains a few more special predicates which make it easier to handle lists.  In our list-extension of $\nthree{}^\exists$ we include\footnote{The list predicates are specified at \url{https://w3c.github.io/N3/reports/20230703/builtins.html\#list}. We exclude the rather complex predicates \texttt{list:iterate} and \texttt{list:memberAt}.} \texttt{list:last}, \texttt{list:in}, \texttt{list:member}, \texttt{list:append}, and \texttt{list:remove}:
\texttt{list:last} is used to relate a list to its last argument\footnote{We give an example of one or more triples (in brackets) which need to be true after each explanation.} ( \texttt{(:a :b :c) list:last :c.}),  \texttt{list:member} defines the relation between a list and its member ( \texttt{(:a :b :c) list:member :a, :b, :c.}), \texttt{list:in} is the inverse of \texttt{list:member} (\texttt{:b list:in (:a :b :c).}), \texttt{list:append} expresses that the list in object position is the combination of the two lists in subject position (\texttt{((:a~:b)~(:c :d)) list:append (:a :b :c :d).}), and by \texttt{list:remove} we express that the object list is the list we get by removing all occurrences of the second argument of the subject list of the first argument of the subject list (\texttt{((:a~:b~:a~:c)~:a)~list:remove ~(:b~:c).}). 

Note, that \nthree built-ins are not defined as functions but as relations.  As a consequence of that, they can be used in different ways. We illustrate this on the predicate \texttt{list:append}. If we write the following rule
\begin{equation}\label{eq:simple-append}
	\texttt{\{((:a :b) (:c :d)) list:append ?x\}=>\{:result :is ?x\}.}
\end{equation}
a reasoner will retrieve
\begin{equation}
	\texttt{:result :is (:a :b :c :d).}
\end{equation}
But we can also write a rule like
\begin{equation}\label{eq:complex-append}
	\texttt{\{(?x ?y) list:append (:a :b :c)\}=>\{?x :and ?y\}.}
\end{equation}
which yields
\begin{align*}
	& \texttt{() :and (:a :b :c).}\\
	& \texttt{(:a) :and (:b :c).}\\
	& \texttt{(:a :b) :and ( :c).}\\
	& \texttt{(:a :b :c) :and ().}
\end{align*}
Additionally, it is possible that only one of the two varaibles in the subject list is instantiated, with
\begin{equation}
	\texttt{\{((:a :b) ?y) list:append (:a :b :c)\}=>\{:we :get ?y\}.}
\end{equation}
for example, we get
\begin{equation}
	\texttt{:we :get (:c).}
\end{equation}
On a practical level, however, this understanding of built-ins as relations comes with some limitations. If the presence of a built-in predicate 
causes a rule to produce infinitely many results, like it is the case with
\begin{equation}
	\texttt{\{?x list:last :c\}=>\{:we :get ?x\}.}
\end{equation}
where all possible lists having \texttt{:c} as last element need to be produced, reasoning engines normally ignore the rule.\footnote{To be more precise, the \nthree specification comes with so-called argument-modes specifying which arguments need to be instatntiated for the predicate to be called, see also \citeauthor{N3-builtins}.}  We will define the full meaning of built-in predicates in our semantics, but our translation to existential rules provided in the next section will only focuss on built-in predicates producing a limited number of solutions.

We now come to the semantics of list predicates. Given an \nthree alphabet which contains the list constants  \texttt{rdf:first}, \texttt{rdf:rest}, \texttt{list:in}, \texttt{list:member}, \texttt{list:append}, \texttt{list:last} and \texttt{list:remove}, and an \nthree Interpretation $\mathfrak{I}=(\mathfrak{D}, \mathfrak{a}, \mathfrak{p})$.
We say that $\mathfrak{I}$ is a model according to list semantics of a formula $\phi$, written as $\mathfrak{I}\models_l \phi$ iff  $\mathfrak{I}\models_{sl}\phi$ and the following conditions hold:
\begin{itemize}
	\item  $\mathfrak{I}\models_l  s\ \texttt{list:in}\ o.$ iff $\mathfrak{a}(o)=(o_1 \ldots o_n)$ and  $\mathfrak{a}(s)= o_i$  for some $i$ with $1 \leq i \leq n$,
		\item  $\mathfrak{I}\models_l  s\ \texttt{list:member}\ o.$ if $\mathfrak{a}(s)=(s_1 \ldots s_n)$ and  $\mathfrak{a}(o)= s_i$  for some $i$ with $1 \leq i \leq n$,
	\item $\mathfrak{I}\models_l  s\ \texttt{list:append}\ o.$ iff $\mathfrak{a}(s)=((a_1 \ldots a_n)(b_1 \ldots b_m))$, $0\leq n$,  $0\leq m$, and   $\mathfrak{a} (o)=(a_1 \ldots a_n \, b_1 \ldots b_m)$,
		\item  $\mathfrak{I}\models_l  s\ \texttt{list:last}\ o$ iff $\mathfrak{a}(s)=(s_1 \ldots s_n)$ and $\mathfrak{a}(o)=s_n$,
			\item  $\mathfrak{I}\models_l  s\ \texttt{list:remove}\ o$ iff $\mathfrak{a}(s)=((a_1 \ldots a_n)\, b)$ and $\mathfrak{a} (o)=(a_i)_{a_i\neq b}$
\end{itemize}
In the next section we discuss how lists and list predicates can be modeled with existential rules.

 \newcommand\listS{\textit{list}\xspace}
\newcommand\emptyS{\textit{empty}\xspace}
\newcommand\consS{\textit{cons}\xspace}
\newcommand\getListS{\textit{getList}\xspace}
\newcommand\getAppendS{\textit{getAppend}\xspace}
\newcommand\getAppendSS{\textit{getAppendS}\xspace}
\newcommand\appS{\textit{append}\xspace}
\newcommand\appsS{\textit{appendS}\xspace}
\newcommand\isFirstS{\textit{isFirst}\xspace}
\newcommand\firstS{\textit{first}\xspace}
\newcommand\restS{\textit{rest}\xspace}
\newcommand\lastS{\textit{last}\xspace}
\newcommand\inS{\textit{isIn}\xspace}
\newcommand\getRemoveS{\textit{getRemove}\xspace}
\newcommand\removeS{\textit{removed}\xspace}
\newcommand\nil{\texttt{rdf:nil}\xspace}
\subsection{Implementing N3 Lists in Exitential Rules}\label{exlist}
We model lists alongside the RDF representation of the previous subsection, sticking to the criteria imposed by N3, predominantly \emph{uniqueness of lists} and \emph{functionality}.
For readability purposes we subsequently diverge from using our triple predicate $\tripl$ for predicates concerning lists.
Instead of $\tripl(x, \texttt{rdf:first}, y)$ we use an auxiliary binary predicate $\firstS$ and write $\firstS(x,y)$.
Similarly we use $\restS(x,y)$ to denote $\tripl(x,\texttt{rdf:rest}, y)$.
For technical reasons, we use a unary predicate $\listS$ to identify all those objects that are lists.
Before modeling lists and their functions, let us formulate the criteria based on the three predicates:
A model $\Inst[M]$ of rule set $\Sigma$ and database $\Dnst$ satisfies
\begin{description}
  \item[Uniqueness of Lists] if for all lists $l_1$ and $l_2$ (i.e., $\listS(l_1), \listS(l_2) \in \Inst[M]$), $\firstS(l_1, x), \firstS(l_2, x)\in\Inst[M]$ and $\restS(l_1, r), \restS(l_2, r) \in \Inst[M]$ implies $l_1 = l_2$;  
  \item[Functionality] if for all lists $l$ (i.e., $\listS(l) \in \Inst[M]$), $\firstS(l, x), \firstS(l, y) \in \Inst[M]$ implies $x=y$, and $\restS(l, x), \restS(l, y)\in \Inst[M]$ implies $x=y$.
\end{description}
Towards \textbf{Existence of Lists}, we ensure existence of the empty list:
\begin{eqnarray}
  \label{eq:emptylist}
  & \rightarrow & \listS(\nil)  \end{eqnarray}
Given that many rule reasoners operate via materialization of derived facts, we should not fully implement the \textbf{Existence of Lists} criterion since materializing all lists certainly entails an infinite process.
Instead, we create lists on-demand.
The binary \getListS predicate expects a list element $x$ (to be added) and a list $l$, and creates a new list with \emph{first} element $x$ and \emph{rest} $l$:
\begin{eqnarray}
  \label{eq:getlist}
  \getListS(x, l) \wedge \listS(l) & \rightarrow & \exists l' .\ \listS(l') \wedge \firstS(l', x) \wedge \restS(l', l)
\end{eqnarray}
With this interface in place, we replicate example~\eqref{list} as follows:
\begin{align*}
  & \rightarrow \getListS(\texttt{:tea}, \nil) \\
  \firstS(l, \texttt{:tea}) \wedge \restS(l, \nil) & \rightarrow \getListS(\texttt{:chocolate}, l) \\
  \firstS(l, \texttt{:chocolate}) \wedge \restS(l, l') \wedge \\
  \firstS(l', \texttt{:tea}) \wedge \restS(l', \nil) & \rightarrow \getListS(\texttt{:tea}, l) \\
  \firstS(l, \texttt{:cake}) \wedge \restS(l, l') \wedge \\
  \firstS(l', \texttt{:chocolate}) \wedge \restS(l', l'') \wedge \\
  \firstS(l'', \texttt{:tea}) \wedge \restS(l'', \nil)& \rightarrow \tripl(\texttt{:lucy}, \texttt{:likes}, l)
\end{align*}
This rather cumbersome encoding implements \textbf{Uniqueness of Lists}.
Towards a much simpler encoding, suppose we only take the following rule obtaining the same list as above:
\begin{equation}
  \begin{array}{rcl}
    & \rightarrow \exists l_1, l_2, l_3 . & \listS(l_1) \wedge \listS(l_2) \wedge \listS(l_3) \wedge \\
  && \firstS(l_1, \texttt{:cake}) \wedge \restS(l_1, l_2) \wedge \\
  && \firstS(l_2, \texttt{:chocolate}) \wedge \restS(l_2, l_3) \wedge \\ 
  && \firstS(l_3, \texttt{:tea}) \wedge \restS(l_3, \nil)
  \end{array}\label{eq:simple-lists}
\end{equation}
The rule itself can now be combined with other rules as well as the previous one.
However, uniqueness can be violated when the restricted chase is used for reasoning.
Recall from \Cref{sub:rule-prelims} that the restricted chase creates new facts (by instantiating rule heads) only if the rule matches are not yet satisfied.
Suppose we create an alternative list that is the same as before but replaces \texttt{:cake} for \texttt{:cookies}:
\begin{equation}
  \begin{array}{rcl}
    & \rightarrow \exists l_1, l_2, l_3 . & \listS(l_1) \wedge \listS(l_2) \wedge \listS(l_3) \wedge \\
  && \firstS(l_1, \texttt{:cookies}) \wedge \restS(l_1, l_2) \wedge \\
  && \firstS(l_2, \texttt{:chocolate}) \wedge \restS(l_2, l_3) \wedge \\ 
  && \firstS(l_3, \texttt{:tea}) \wedge \restS(l_3, \nil)
  \end{array}\label{eq:uniqueness-counterexample}
\end{equation}
While the list created by rule~\eqref{eq:uniqueness-counterexample} is surely distinct from the one created through rule application of \eqref{eq:simple-lists}, they also obtain different sublists.
After a restricted chase over rule set $\{ \eqref{eq:simple-lists}, \eqref{eq:uniqueness-counterexample} \}$ and the empty database, we get two distinct lists $l$ and $l'$ such that $\firstS(l, \texttt{:tea})$, $\firstS(l', \texttt{:tea}), \restS(l,\nil), \restS(l',\nil)$, contradicting \textbf{Uniqueness of Lists}.
The reason for this is that the application condition of the restricted chase checks whether the head of the rule is already satisfied.
If not, the full head is instantiated with (globally) fresh nulls in place of the existentially quantified variables.
Our encoding via rule~\eqref{eq:getlist} overcomes this issue by step-wise introducing new list elements.
If a sublist already exists, rule creation is not triggered redundantly.

\begin{theorem}\label{prop:uniqueFunctionalExistLists}
  Let \Dnst be a database, $\Sigma$ a rule set, and $\Inst$ the restricted chase of $\Sigma$ and $\Dnst$.
  If the only rules in $\Sigma$ using predicates \listS, \firstS, or \restS in their heads are those of \eqref{eq:emptylist} and \eqref{eq:getlist}, then \Inst satisfies (a) \textbf{Uniqueness of Lists} and (b) \textbf{Functionality}.
\end{theorem}
\begin{proof}
  \textbf{Functionality} follows from the fact that the only rule introducing \firstS- and \restS-atoms is \eqref{eq:getlist} and, thereby, uniquely determines first and rest elements for a list term.
  Thus, predicates \firstS and \restS are functional.

  Regarding \textbf{Uniqueness of Lists}, we observe that only rule \eqref{eq:getlist} introduces lists together with their (functional) \firstS and \restS atoms.
  Hence, if there were two lists $l_1$ and $l_2$ with the same first and rest elements, then the respective chase sequence $\Dnst^0 \Dnst^1 \Dnst^2 \ldots$ contains a member $\Dnst^i$ in which (without loss of generality) $l_1$ is contained.
  Furthermore, there is a later instance $\Dnst^j$ ($j>i$) in which $l_2$ is not yet contained but is about to be added to $\Dnst^{j+1}$.
  But rule \eqref{eq:getlist} is already satisfied in $\Dnst^j$ for the respective first/rest elements.
  Thus, $l_2$ will never be instantiated by the restricted chase and can, thus, not be part of the chase.
\end{proof}
Before we get into the intricates of appending two or more lists, let us briefly show the rules for implementing \texttt{list:last} and \texttt{list:in} (and \texttt{list:member} as the inverse of \texttt{list:in}), represented by binary predicate symbols \lastS and \inS.
\begin{eqnarray}
  \firstS(l, x) \wedge \restS(l, \nil) & \rightarrow & \lastS(x, l) \\
  \restS(l, l') \wedge \lastS(y, l') & \rightarrow & \lastS(y, l) \\
  \firstS(l, x) & \rightarrow & \inS(l, x) \\
  \restS(l, l') \wedge \inS(l', y) & \rightarrow & \inS(l, y) 
\end{eqnarray}
Note, these rules are sufficient for creating all necessary facts to obtain the required results.
Regarding list concatenation via \texttt{list:append}, we introduce the ternary predicate \appS with the appended list in the first position and the two constituent lists in second and last.
First, every list $l$ \emph{prepended} by the empty list yields itself:
\begin{eqnarray}\label{eq:append-nil}
  \listS(l) \rightarrow \appS(l, \nil, l)
\end{eqnarray}
Second, if we append lists $l_1$ and $l_2$ to get $l_3$ (i.e., $\appS(l_3, l_1, l_2)$), and $x$ is the first element of $l_2$, then $l_3$ can also be obtained by appending $x$ to $l_1$, and the result to the rest of $l_2$.
Therefore, we need an auxiliary set of rules that appends a single element $x$ to a list $l$:
\begin{eqnarray}
  \label{eq:append-single}\appS(l_3, l_1, l_2) \wedge \firstS(l_2, x) & \rightarrow & \getAppendSS(l_1, x) \\
  \label{eq:append-single-rec}\getAppendSS(l, x) \wedge \restS(l, l') & \rightarrow & \getAppendSS(l', x)
\end{eqnarray}
Rule~\eqref{eq:append-single} requests a new list that starts with the same elements as $l_1$ and appends the additional element $x$.
Rule~\eqref{eq:append-single-rec} recursively pushes the request through the list.
Once, the empty list (\nil) is reached, appending the element $x$ is the same as prepending it to \nil:
\begin{equation}
  \begin{array}{rcl}
  \getAppendSS(\nil, x) & \rightarrow & \getListS(x, \nil) \\
  \getAppendSS(\nil, x) \wedge \listS(l) \wedge \\
  \firstS(l, x) \wedge \restS(l, \nil) & \rightarrow & \appsS(l, \nil, x) \\
  \end{array}
\end{equation}
These rules create a fresh list with first element $x$ and rest \nil if necessary.
Predicate \appsS stands for \emph{append singleton} and, therefore, $\appsS(l, l', x)$ tells that list $l$ is the result of appending $x$ to list $l'$.
The recursive step is implemented as follows:
\begin{equation}\label{eq:append-singleton}
  \begin{array}{rcl}
  \getAppendSS(l, x) \wedge \firstS(l, y) \wedge \restS(l, l') \wedge \appsS(l'', l', x) & \rightarrow & \getListS(y, l'') \\
  \getAppendSS(l, x) \wedge \firstS(l, y) \wedge \restS(l, l') \wedge \\
  \appsS(l'', l', x) \wedge
  \listS(l_{\nu}) \wedge \firstS(l_{\nu}, y) \wedge \restS(l_{\nu}, l'')  & \rightarrow & \appsS(l_{\nu}, l, x)
  \end{array}
\end{equation}
So if a list $l$ shall be appended by singleton $x$ and we already know that for the rest of $l$ (i.e., $l'$) there is a version with appended $x$ (i.e., $l''$), then $l$ appended by $x$ is the new list formed by the first element of $l$ (i.e., $y$) and $l''$ as rest.

Last, appending two lists can also be requested via rules.
Once more, we use a predicate for this request, namely \getAppendS.
This predicate is an interface for users (i.e., other rules) to create lists beyond predicate \getListS.
Such requests are served by the following rules:
\begin{eqnarray}
  \getAppendS(\nil, l_2) & \rightarrow & \appS(l_2, \nil, l_2) \\
  \getAppendS(l_1, l_2) \wedge \firstS(l_1, x) \wedge \restS(l_1, l_1') & \rightarrow & \getAppendS(l_1', l_2) \\
  \getAppendS(l_1, l_2) \wedge \firstS(l_1, x) \wedge \restS(l_1, l_1') \wedge \\ \wedge \appS(l_3,l_1', l_2) & \rightarrow & \getListS(x, l_3) \\
  \getAppendS(l_1, l_2) \wedge \firstS(l_1, x) \wedge \restS(l_1, l_1') \wedge \\  \appS(l_3,l_1', l_2) \wedge \firstS(l_3', x) \wedge \restS(l_3', l_3) & \rightarrow & \appS(l_3', l_1, l_2)
\end{eqnarray}

The remove functionality can be implemented in a similar fashion.
Note that none of the additionally instantiated rules for list built-ins use predicates \listS, \firstS, or \restS in their heads.
Thus, \Cref{prop:uniqueFunctionalExistLists} still holds in rule sets using built-in functions.
Throughout the rest of this subsection we aim at showing how the framework implements the examples given throughout \Cref{sub:n3lists} as well as an example of list usage inside \nthree rules.

\paragraph{Appending Lists.}
First, recall the following \nthree rule (cf.\ \eqref{eq:simple-append}):
\begin{equation*}
	\texttt{\{((:a :b) (:c :d)) list:append ?x\}=>\{:result :is ?x\}.}
\end{equation*}
For the implementation of this rule, we need to make sure the constant lists (the operands of \texttt{list:append}) exist:
\begin{eqnarray*}
  & \rightarrow & \getListS(\texttt{:b}, \nil) \\
  \listS(l) \wedge \firstS(l, \texttt{:b}) \wedge \restS(l, \nil) & \rightarrow & \getListS(\texttt{:a}, l) \\
  & \rightarrow & \getListS(\texttt{:d}, \nil) \\
  \listS(l) \wedge \firstS(l, \texttt{:d}) \wedge \restS(l, \nil) & \rightarrow & \getListS(\texttt{:c}, l)
\end{eqnarray*}
After these rules have been used, the lists in example \eqref{eq:simple-append} are guaranteed to exist.
Next, we can request to append the two lists matched within the rule:
\begin{eqnarray*}
  \listS(l_1) \wedge \firstS(l_1, \texttt{:a}) \wedge \restS(l_1, l_1') \wedge && \\
  \firstS(l_1', \texttt{:b}) \wedge \restS(l_1', \nil) \wedge && \\
  \listS(l_2) \wedge \firstS(l_2, \texttt{:c}) \wedge \restS(l_2, l_2') \wedge && \\
  \firstS(l_2',\texttt{:d}) \wedge \restS(l_2', \nil) & \rightarrow & \getAppendS(l_1, l_2)
\end{eqnarray*}
After this rule we are guaranteed to have all lists in place for implementing our rule.
\begin{eqnarray*}
  \listS(l_1) \wedge \firstS(l_1, \texttt{:a}) \wedge \restS(l_1, l_1') \wedge && \\
  \firstS(l_1', \texttt{:b}) \wedge \restS(l_1', \nil) \wedge && \\
  \listS(l_2) \wedge \firstS(l_2, \texttt{:c}) \wedge \restS(l_2, l_2') \wedge && \\
  \firstS(l_2',\texttt{:d}) \wedge \restS(l_2', \nil) \wedge && \\
  \appS(x, l_1, l_2) & \rightarrow & \tripl(\texttt{:result}, \texttt{:is}, x)
\end{eqnarray*}
Second, we reconsider rule~\eqref{eq:complex-append}:
\begin{equation*}
	\texttt{\{(?x ?y) list:append (:a :b :c)\}=>\{?x :and ?y\}.}
\end{equation*}
In this example we need to ensure the resulting list exists.
Our rule framework (especially rules \eqref{eq:append-nil}--\eqref{eq:append-singleton}) takes care of disecting the list into its fragment.
Thus, the example rule can be implemented, once the list \texttt{(:a :b :c)} has been created as before, by
\begin{eqnarray*}
  \listS(l) \wedge \firstS(l, \texttt{:a}) \wedge \restS(l, l') \wedge && \\
  \firstS(l', \texttt{:b}) \wedge \restS(l', l'') \wedge && \\
  \firstS(l'', \texttt{:c}) \wedge \restS(l'', \nil) \wedge && \\
  \appS(l, x, y) & \rightarrow & \tripl(x, \texttt{:and}, y)
\end{eqnarray*}

\paragraph{List Creation in Rules.}
Last, we consider an \nthree rule that identifies two lists in its body and creates a new list based on some elements identified within the list.
The following rule identifies two lists, one with three elements (\texttt{?x}, \texttt{?y}, and \texttt{?z}) and one with two elements (\texttt{?a} and \texttt{?b}), and then creates a new list with first element \texttt{?y} and a rest list with the singleton element \texttt{?b}:
\begin{equation}\label{eq:n3rules}
  \texttt{\{:s :p (?x ?y ?z). :k :l (?a ?b)\}=>\{:h :i (?y ?b)\}.}
\end{equation}
This rule needs splitting into creating the list for the result and then creating the output triple:
\begin{eqnarray*}
  \listS(l_1) \wedge \firstS(l_1, x) \wedge \restS(l_1, x_l) \wedge \\
  \firstS(x_l, y) \wedge \restS(x_l, y_l) \wedge \\
  \firstS(y_l, z) \wedge \restS(y_l, \nil) \wedge \\
  \listS(l_2) \wedge \firstS(l_2, a) \wedge \restS(l_2, a_l) \wedge \\
  \firstS(a_l, b) \wedge \restS(a_l, \nil) \wedge \\
  \tripl(\texttt{:s}, \texttt{:p}, l_1) \wedge \tripl(\texttt{:k}, \texttt{:l}, l_2) & \rightarrow & \getListS(b, \nil) \\
\end{eqnarray*}
\begin{eqnarray*}
  \listS(l_1) \wedge \firstS(l_1, x) \wedge \restS(l_1, x_l) \wedge \\
  \firstS(x_l, y) \wedge \restS(x_l, y_l) \wedge \\
  \firstS(y_l, z) \wedge \restS(y_l, \nil) \wedge \\
  \listS(l_2) \wedge \firstS(l_2, a) \wedge \restS(l_2, a_l) \wedge \\
  \firstS(a_l, b) \wedge \restS(a_l, \nil) \wedge \\
  \tripl(\texttt{:s}, \texttt{:p}, l_1) \wedge \tripl(\texttt{:k}, \texttt{:l}, l_2) \wedge \\
  \listS(l) \wedge \firstS(l, b) \wedge \restS(l, \nil) & \rightarrow & \getListS(y, l) \\
\end{eqnarray*}
\begin{eqnarray*}
  \listS(l_1) \wedge \firstS(l_1, x) \wedge \restS(l_1, x_l) \wedge \\
  \firstS(x_l, y) \wedge \restS(x_l, y_l) \wedge \\
  \firstS(y_l, z) \wedge \restS(y_l, \nil) \wedge \\
  \listS(l_2) \wedge \firstS(l_2, a) \wedge \restS(l_2, a_l) \wedge \\
  \firstS(a_l, b) \wedge \restS(a_l, \nil) \wedge \\
  \tripl(\texttt{:s}, \texttt{:p}, l_1) \wedge \tripl(\texttt{:k}, \texttt{:l}, l_2) \wedge \\
  \listS(l') \wedge \firstS(l', b) \wedge \restS(l', \nil) \wedge \\
  \listS(l) \wedge \firstS(l, y) \wedge \restS(l, l') & \rightarrow & \tripl(\texttt{:h}, \texttt{:i}, l)
\end{eqnarray*}
This construction may become complicated if several list built-ins are co-dependent.
 \subsection{\nthree List Predicates as Syntactic Sugar}
\label{n3listrules}
As detailed in the previous section, \nthree list predicates can be expressed by means of existential rules if the reasoning is performed under similar premises as the restricted chase. This is particularly interesting in the context of Notation3 Logic: it is well-known that list predicates \texttt{list:in},  \texttt{list:member}, \texttt{list:append}, \texttt{list:last}, and \texttt{list:remove} introduced in \Cref{sub:n3lists} are syntactic sugar, and, therefore, can be expressed using rules in combination with the predicates \texttt{rdf:first} and \texttt{rdf:rest}. Typically these rules are only written for reasoners supporting backward-chaining, that is, with algorithms performing reasoning starting from the goal and following rules from head to body until some factual evidence is found\footnote{This kind of reasoning is very similar to Prolog's SLD resolution \citep{nilsson}.}.

For better illustration, consider the following \nthree rules implementing \texttt{list:append}\footnote{\nthree allows rules to be written in a backwards, that is instead of \texttt{A=>B.} we write \texttt{B<=A.} The backward notation is usally used to indicate that this rule is expected to be reasoned with via backward-chaining. We use this notation here, the model-theoretic semantics keeps being the same as before.}:
\begin{align}
		\texttt{\{(() ?x) list:append ?x\}<=}& \texttt{\{ \}. }\label{ap1}\\
	\texttt{\{(?x ?y) list:append ?z\}<=}& \texttt{\{?x rdf:first ?a. ?x rdf:rest ?r.}\label{ap2}\\ &\texttt{ ?z rdf:first ?a. ?z rdf:rest ?q. }\notag \\ &\texttt{ (?r ?y) list:append ?q \}  }.\notag
\end{align}
If these rules are used in backward-chaining, they get triggered by each execution of a rule containing a triple with the predicate \texttt{list:append}. If we, for example, would like to get all instances of the triple \texttt{:result :is ?x.} which can be derived by rule
\eqref{eq:simple-append}, the triple in the body of the rule triggers rule \eqref{ap2}, to test whether there is evidence for the triple
\texttt{((:a :b) (:c :d)) list:append ?x.} The rule is again followed in a backwards direction yielding:
\begin{align}
&	\texttt{ (:a :b) rdf:first :a; rdf:rest (:b).}\label{oben1}\\ &\texttt{ ?x rdf:first :a;  rdf:rest ?q. }\notag \\ &\texttt{ ((:b) (:c :d)) list:append ?q.  }.\notag
\end{align}
The triples in the first line of this example got instantiated according to the semantics of \texttt{rdf:first} and \texttt{rdf:rest}. This istantiation also caused the triples in the following two lines to partially instantiated. Since there is not enough information to instantiate the triples from the second line, a (backward) reasoner would continue with the last triple which again has \texttt{list:append} in predicate position.
Rule \eqref{ap2} is called again. This time we retrieve:
\begin{align}
	&	\texttt{ (:b) rdf:first :b; rdf:rest ().}\label{oben2}\\ &\texttt{ ?q rdf:first :b;  rdf:rest ?q2. } \notag \\ &\texttt{ (() (:c :d)) list:append ?q2.  }.\notag
\end{align}
Again following the rules backwards, we can apply rule \eqref{ap1} to get a value for \texttt{?q2}:
\[
\texttt{ (() (:c :d)) list:append (:c :d).  }
\]
With this information, we get a binding for \texttt{?q} in \eqref{oben2}:
\begin{align*}
	&	\texttt{ (:b) rdf:first :b; rdf:rest ().}\label{exe1}\\ &\texttt{ (:b :c :d) rdf:first :b;  rdf:rest (:c :d). }\notag \\ &\texttt{ (() (:c :d)) list:append (:c :d).  }.\notag
\end{align*}
Subsequently, we obtain a new binding \texttt{?x} in \eqref{oben1}:
\begin{align*}
	&	\texttt{ (:a :b) rdf:first :a; rdf:rest (:b).}\\ &\texttt{ (:a :b :c :d) rdf:first :a;  rdf:rest  (:b :c :d ). } \\ &\texttt{ ((:b) (:c :d)) list:append (:b :c :d).  }.
\end{align*}
This produces \texttt{:we :get (:a :b :c :d).} as a solution. The backward-chaining process produces triples on-demand: only if a rule premise depends on the information, a backward rule is called to retrieve it, and this allows us to have infinitely large models which we do not materialize during reasoning. 

In the \nthree community, this and other examples are normally used to argue that \nthree reasoners should support backwards-reasoning as a way to only produce triples when these are needed to find instances for a goal.  Following the findings of the previous subsection, it is not true that we necessarily need backward rules to support triple production on-demand. Instaed of writing rule \eqref{ap1} and \eqref{ap2}, we can also add the triple 
\texttt{(:a~:b)~:getAppend~(:c :d).}
to our initial rule \eqref{eq:simple-append}. With the following rules, we retrieve the same result as above:
\begin{verbatim}
{() :getAppend ?y}}=>{(() ?y) list:append ?y}.
{?x :getAppend ?y; rdf:rest ?b}=>{?b :getAppend ?y}.
{?x :getAppend ?y; rdf:first ?a; rdf:rest ?b.
 (?b ?y) list:append ?z. ?z2 rdf:first ?a ; rdf:rest ?z }
	  =>{(?x ?y) list:append ?z2 }.}
\end{verbatim}
These rules follow the structure of the rules in the previous subsection with the exception that we do not need list constructors in \nthree. If we apply our rules to the fact above, we  sucessively construct the triples \texttt{(() (:c :d)) list:append (:c :d).}, 
\texttt{((:b) (:c :d)) list:append (:b :c :d).}, and \texttt{((:a :b) (:c :d)) list:append (:a :b :c :d).}. These can then directly be used in rules. In more complicated cases, where the arguments of the predicate \texttt{list:append} 
do not appear partially instantiated in rule bodies, the relevant instances of the fact \texttt{?x :getAppend ?y.} need to be constructed via rules just as it is the case for existential rules. As \nthree follows the axioms introduced in \Cref{sub:n3lists}, the first-rest interpretation of RDF lists is equilvalent to \nthree{}'s representation of lists as first-class citizens. As a consequence, the rules actually work for all examples introduced above. Similarly, the other list predicates can be written by means of \texttt{rdf:first} and \texttt{rdf:rest}, and handled via backward-chaining or, alternatively, with some version of the chase.

Note, the backward rules handling \texttt{list:append} can be mimicked by splitting them in several forward rules acting on a \emph{getter triple}, that is, a triple causing the production of the required instance of the predicate. We additionally need rules producing the required instances of that getter triple, at which point we need to be careful considering dependencies between triples. However, the mechanism introduced in the previuos subsection provides us with a possibility to do reasoning on-demand in a purely forward manner.
  \section{Evaluation}\label{imp}
The considerations provided above allow us to use existential rule reasoners to perform $\nthree{}^\exists$ reasoning.
We would like 
to find out whether  our finding is of practical relevance, that is
whether we can identify datasets on which
existential rule reasoners, running on the rule translations,
outperform classical \nthree reasoners provided with the original data.

In order to do this we have 
implemented $\transf$ as a python
script that takes an arbitrary \nex formula $f$, constructs its set representation $F$ in PNF, and produces the set of rules $\transf(F)$.
This script and some additional scripts to translate existential rules (with at most binary predicates) to \nex formulae are available on \href{https://github.com/smennicke/n32rules}{GitHub}.
Our implementation allows us to compare \nthree reasoners with existential rule reasoners, performance-wise.
As existential rule reasoners we chose VLog \citep{VLog4j2019}, a state-of-the-art reasoning engine designed for working with large piles of input data, and Nemo \citep{nemo23}, a recently released rust-based reasoning engine.
As \nthree reasoners we chose cwm \citep{cwm} and EYE \citep{eyepaper} which -- due to their good coverage of \nthree features -- are most commonly used.
All experiments have been performed on a laptop with 11th Gen Intel Core i7-1165G7 CPU, 32GB of RAM, and 1TB disk capacity, running a Ubuntu 22.04 LTS.

\subsection{Datasets}\label{sub:datasets}
We performed our experiments on two datasets: \textsc{Lubm} from the \emph{Chasebench} \citep{Benedikt+17:ChaseBench} provides a fixed set of 136 rules and 
varies in the number of facts these rules are applied; the \textsc{Deep Taxonomy} (DT) benchmark developed for the \emph{WellnessRules} project \citep{BoleyOC09:WellnessRules}  consists of one single fact and a varying number of mutually dependent rules.

The \emph{Chasebench} is a benchmarking suite for existential rule reasoning.
Among the different scenaria in Chasebench we picked \textsc{Lubm} for its direct compatibility with \nthree: all predicates in \textsc{Lubm} have at most arity $2$. Furthermore, \textsc{Lubm} allows for a glimpse on scalability since \textsc{Lubm} comes in different database sizes.
We have worked with \textsc{Lubm} 001, 010, and 100, roughly referring to dataset sizes of a hundred thousand, one million and ten million facts.
We translated \textsc{Lubm} data and rules into a canonical \nthree format.
Predicate names and constants within the dataset become IRIs using the example prefix.
An atom like $\textit{src\_advisor}(\textit{Student441},\textit{Professor8})$ becomes the triple $\ntt{:Student441 :src\_advisor :Professor8.}$.
For atoms using unary predicates, like $\textit{TeachingAssistent}(\textit{Student498})$, we treat $\ntt{:TeachingAssistent}$ as a class and relate $\ntt{:Student498}$ via $\ntt{rdf:type}$ to the class.
For any atom $A$, we denote its canonical translation into triple format by $t(A)$.
Note this canonical translation only applies to atoms of unary and binary predicates.
For the existential rule\\[.3em]
\centerline{$\forall \vec x .\ B_{1} \wedge \ldots \wedge B_{m} \rightarrow \exists \vec z .\ H_{1} \wedge \ldots \wedge H_{n}$}\\[.3em]
we obtain the canonical translation by applying $t$ to all atoms, respecting universally and existentially quantified variables (i.e., universally quantified variables are translated to universal \nthree variables and existentially quantified variables become blank nodes):\\[.3em]
\centerline{$\ntt\{ t(B_{1})\ntt. \cdots t(B_{m})\ntt. \ntt\}\ntt{=>}\ntt\{ t(H_{1})\ntt. \cdots t(H_{n})\ntt. \ntt\}\text.$}
All \nthree reasoners have reasoned over the canonical translation  of data and rules which was necessary because of the lack of an \nthree version of \textsc{Lubm}. Since we are evaluating VLog's and Nemo's performance on our translation $\transf$, we converted the translated \textsc{Lubm} by $\transf$ back to existential rules before reasoning.
Thereby, former unary and binary atoms were turned into triples and then uniformly translated to $\textit{tr}$-atoms via $\transf$. 

\begin{figure}[tbp]
\centering
    \footnotesize
\begin{tikzpicture}
	\node[object] (x) at (-2,0) {x} ;
	\node[object] (N0) at (0,0) {$N_0$} ;
	\node[object] (N1) at (3.5,0) {$N_1$} ;
	\node[object] (I1) at (3.5,-0.8) {$I_1$} ;
	\node[object] (J1) at (3.5,-1.6) {$J_1$} ;
	\node[object] (N2) at (7,0) {$N_2$} ;
		\node[object] (I2) at (7,-0.8) {$I_2$} ;
			\node[object] (J2) at (7,-1.6) {$J_2$} ;
				\node (N3) at (8,0) {...} ;
			\node (I3) at (8,-0.4) {...}  ;
			\node (J3) at (8,-0.8) {...} ;
				\draw[->] (x) --node[predicate] {a} (N0) ;
	\draw[->] (N0) --node[predicate] {subClassOf} (N1) ;
			\draw[->] (N0) --node[predicate] {subClassOf} (J1);
					\draw[->] (N0) --node[predicate] {subClassOf} (I1);
	\draw[->] (N1) --node[predicate] {subClassOf} (N2) ;
			\draw[->] (N1) --node[predicate] {subClassfOf} (J2);
\draw[->] (N1) --node[predicate] {subClassOf} (I2);
	\draw (N2) -- (N3);
	\draw (N2) -- (I3);
		\draw (N2) -- (J3);
\end{tikzpicture}
\normalsize
\caption{Structure of the  \textsc{Deep Taxonomy} benchmark. \label{ws}}
\end{figure}

The \emph{Deep Taxonomy benchmark}
simulates deeply nested RDFS-subclass reasoning\footnote{\nthree available at: \url{http://eulersharp.sourceforge.net/2009/12dtb/}.}. It contains one individual which is member of a class.
This class is subclass of three other classes of which one again is subclass of three more classes and so on. \Cref{ws} illustrates this idea. The benchmark provides different  depths for the subclass chain and we tested with the depths of 1,000 and 100,000. The reasoning tests for the membership of the individual in the last class of the chain.  
For our tests,
 the subclass declarations were translated  to rules, the triple 
$\texttt{:N0 rdfs:subClassOf :N1.}$
became\\[.3em]
\centerline{\texttt{\{ ?x a :N0.\}=>\{ ?x a :N1.\}.}}\\[.3em]
This translation also illustrates why this rather simple reasoning case is interesting: we have a use case in which we depend on long chains of rules executed after each other. The reasoner EYE allows the user to decide per  rule 
whether it is applied using forward- or backward-reasoning, at least if the head of the rule does not contain blank nodes. For this dataset, we evaluated full backward- and full forward-reasoning, separately.

\subsection{Results}\label{sub:results}
\Cref{tab:runtime}  presents the running times of the four reasoners and additionally gives statistics about the sizes of the given knowledge base (\# facts) and the rule set (\# rules).  For DT we display two reasoning times for EYE, one produced by only forward reasoning (EYE-fw), one for only backward-reasoning (EYE-bw). Note, that for the latter, the reasoner does not produce the full deductive closure of the dataset, but answers a query instead. As \textsc{Lubm} contains rules with blank nodes in their haeds, full backward reasoning was not possible in that case, the table is left blank.
EYE performs much better than VLog and Nemo for the experiments with DT. Its reasoning time is off by one order of magnitude.
Conversely, VLog and Nemo could reason over all the \textsc{Lubm} datasets 
 while EYE has thrown an exception after having read the input facts.  
 The reasoning times of VLog are additionally significantly lower than the times for EYE.
 While Nemo shows a similar runtime on DT as VLog, it is slower on \textsc{Lubm}.
 However, we may be quite optimistic regarding its progress in runtime behavior, as Nemo already shows better running times on the original \textsc{Lubm} datasets.
The reasoner cwm is consistently slower than the other three and from \textsc{Lubm 010} on.
All reasoners tried to find the query answers/deductive closures for at least ten minutes (i.e., --- in \Cref{tab:runtime} indicates a time-out).

\begin{table}[tbp]
  \caption{Experimental Results }
  \label{tab:runtime}
  \centering
  \begin{tabular}{lrrrrrrrrr}
    \toprule
    \textbf{Dataset} & \textbf{\# facts} & \textbf{\# rules} & \textbf{cwm} & \textbf{EYE-fw} & \textbf{EYE-bw} & \textbf{VLog} & \textbf{Nemo} \\
    \midrule
    \textsc{DT 1000} & \num{1}     & \num{3001} & \SI{180}{\s}   & \SI{0.1}{\s} & \SI{0.001}{\s} & \SI{1.6}{\s} & \SI{1.7}{\s}  \\
    \textsc{DT 100000} & \num{1}   & \num{30001} & --- & \SI{0.3}{\s} & \SI{0.003}{\s} & --- & --- \\
    \textsc{Lubm 001}      & \num{100543}   & \num{136}  & \SI{117.4}{\s} & \SI{3.4}{\s}  &  & \SI{0.2}{\s} & \SI{2.4}{\s}  \\
    \textsc{Lubm 010}      & \num{1272575}  & \num{136}  & ---            & \SI{44.8}{\s} &  & \SI{4.3}{\s} & \SI{31.2}{\s} \\
    \textsc{Lubm 100}      & \num{13405381} & \num{136}  & ---            & ---           &  & \SI{47.3}{\s} & \SI{362}{\s} \\
    \bottomrule
  \end{tabular}
\end{table}

\subsection{Discussion}\label{sub:discussion}
In all our tests we observe a very poor performance of cwm which is not surprising, given that this reasoner has not been updated for some time. The results for EYE, VLog and Nemo are more interesting as they illustrate the different strengths of the reasoners.

For very high numbers of rules compared to the amount of data, EYE performs much better than
VLog and Nemo. The good results of 0.1  and 0.3 seconds can even be improved by using backward reasoning.
This makes EYE very well-suited for use cases where we need to apply complex rules on datasets of low or medium size. 
This could  be interesting in decentralized set-ups
such as policy-based access control for the Solidproject.\footnote{\url{https://solidproject.org/}.} 
On the other hand we see that VLog and Nemo perform best when provided with large datasets and lower numbers of rules.
This could be useful use cases involving bigger datasets in the Web like
Wikidata or  DBpedia\footnote{\url{https://www.wikidata.org/} \emph{and} \url{https://www.dbpedia.org/}}.

From the perspective of this paper, these two findings together show the relevance of our work: we observed big differences between the tools' reasoning times and these differences  depended on the use cases.
In other words, there are use cases which could benefit from our translation and we thus do not only make the first steps towards having more \nthree reasoners available but also broaden the scope of possible  \nthree applications.

 \section{Related work}\label{relwork}

When originally proposed as a W3C member submission \citep{BernersLeeC11:n3team}, the formal semantics of \nthree was only introduced informally.
As a consequence, different systems, using \nthree, interpreted concepts like nested formulae differently \citep{Arndt19:implicitExplicit}.
Since then, the relation of \nthree to other Web standards has been studied from a use-case perspective \citep{Arndt19:PhD} and a W3C Community group has been formed \citep{N3-spec}, which recently published the semantics of \nthree without functions \citep{N3-semantics}.
Even with these definitions, the semantic relation of the logic to other standards, especially outside the Semantics Web, has not been studied thoroughly.

For \nthree{}'s subset RDF, \citeauthor{erdf} provide a translation to first-order logic and F-Logic using similar
embeddings (e.g., a tenary predicate to represent triples) to the ones in this paper, but do not cover rules.
\citeauthor{Boley16:hub} supports \nthree in his RuleML Knowledge-Interoperation Hub providing a translation of \nthree
to PSOA RuleML. 
This can be translated to other logics. But the focus is more on syntax than on semantics.

In Description Logics (DL), rewritings in rule-based languages
have their own tradition (see, e.g., \citeauthor{CK2020:rewriting-alchiq} for a good overview of existing rewritings and their complexity, as well as more references).
The goal there is to (1) make state-of-the-art rule reasoners available for DLs and, thereby, (2) use a fragment of a rule language that reflects on the data complexity of the given DL fragment.
Also practical tools have been designed to capture certain profiles of the Web Ontology Language (OWL), like the Orel system \citep{KMR10:Orel} and, more recently, DaRLing \citep{darling2020}.
To the best of our knowledge,  a rewriting for \nthree as presented in this paper did not exist before.
Also, existential rule reasoning engines have not been compared to the existing \nthree reasoners.

 \section{Conclusion}\label{conc}

In this paper we studied the close relationship between \nthree rules supporting blank node production and existential rules. 
\nthree without special features like built-in functions, nesting of rules, or quotation can be directly mapped to existential rules with ternary predicates. 
In order to show that, we defined a mapping between \nex -- \nthree without the aforementioned features -- and existential rules.
We argued that this mapping and its inverse preserve the equivalence and non-equivalence between datasets.
This result allows us to trust the reasoning results when applying the mapping in practice, that is, when (1) translating \nex to existential rules, (2) reasoning within that framework, and (3) using the inverse mapping to transfer the result back into \nthree.

We applied that strategy and compared the reasoning times of the \nthree reasoners cwm and EYE with the existential rule
reasoners VLog and Nemo. The goal of that comparison was to find out whether there are use cases for which \nthree reasoning can benefit from the findings on existential rules. We tested the reasoners on two datasets: DT 
consisting of one single fact and a varying number of mutually dependent rules
and \textsc{Lubm} consisting of a fixed number of rules and a varying  number of facts. EYE performs better on DT while VLog and Nemo
showed their strength on \textsc{Lubm}. We see that as an indication that for use cases of similar nature, that is, reasoning 
on large numbers of facts, our approach could be used to improve reasoning times. More generally, we see that reasoners 
differ in their strengths and that, by providing the reversible translation between \nex and existential rules,
we increase the number of  reasoners (partly) supporting \nthree and the range of use cases the logic can support in practice. 
We see our work  as an important step towards fully establishing rule-based reasoning in the Semantic Web.

Of course, \nthree also contains constructs and built-in predicates which are not supported (yet) by our translation.
In order to test how extensible our framework is, we provided strategies to also cover lists and their built-in predicates in the translation.
Lists were constructed using nulls, which made reasoning with them dependent on the chase applicable. 
We provided rules to mimic the list-append function of \nthree under the standard chase, which is also implemented in some \nthree reasoners.
The existential rules version of the append function came with rules that allow for list construction \emph{on-demand}.
This \emph{on-demand}ness is very interesting in many situations and, maybe even more important, believed by the \nthree community to only be possible employing backward reasoning.
In that sense we also contribute to the ongoing discussion in that community whether the intended reasoning direction should be part of the semantics, which we would clearly argue against.

As many \nthree use cases rely on more powerful \nthree{} predicates and logical features such as support for graph terms and nested rules, 
future work should include the extension of our translation towards full coverage of \nthree.
As a direct candidate, we would like to investigate the intricate consequences of non-monotonic reasoning in the presence of existentially quantified variables \citep{EKM22:ExistentialsNonMonotone}.
Another direction of future work is to investigate the differences
and similarities we found in our evaluation in more detail: while showing differences in their performance, the reasoners produced the exact same result sets (modulo isomorphism) when acting on rules introducing blank nodes. 
That is, the different reasoning times do not stem from handling of existentially quantified rule heads but from other optimization techniques.
Fully understanding these differences will help the \nthree and the existential rules communities to further improve their tools.
In that context, it would also be interesting to learn if EYE's capability to combine forward and backward reasoning could improve the reasoning times for data sets including existentially quantified rule heads.

We thus hope that our research on existential \nthree will spawn further investigations of powerful data-centric features in data-intensive rule reasoning as well as significant progress in tool support towards these features.
Ultimately, we envision a Web of data and rule exchange, fully supported by the best tools available as converging efforts of the \nthree community, the existential rule reasoning community, and possibly many others.

\bibliographystyle{acmtrans}

\end{document}